\documentclass[fleqn,10pt]{wlscirep}
\usepackage[utf8]{inputenc}
\usepackage[T1]{fontenc}
\usepackage{graphicx}

\usepackage{tikz}
\usepackage{comment}
\usepackage{amsmath,amssymb} 
\usepackage{enumitem}
\usepackage{booktabs}
\usepackage{cite}
\usepackage{color}
\usepackage{subcaption}
\usepackage[ruled,vlined]{algorithm2e}
\usepackage{breqn}
\usepackage{flushend}
\usepackage[autostyle]{csquotes}
\usepackage{enumitem}
\usepackage{subcaption}
\usepackage{multirow}
\DeclareMathOperator*{\argmax}{arg\,max}

\title{Motion Estimation for Large Displacements and Deformations}

\author[1]{Chen Qiao}
\author[1,*]{Charalambos Poullis}
\affil[1]{Immersive and Creative Technologies Lab, Concordia University, Canada}

\affil[*]{charalambos@poullis.org}

\begin{abstract}
Large displacement optical flow is an integral part of many computer vision tasks. Variational optical flow techniques based on a coarse-to-fine scheme interpolate sparse matches and locally optimize an energy model conditioned on colour, gradient and smoothness, making them sensitive to noise in the sparse matches, deformations, and arbitrarily large displacements. This paper addresses this problem and presents HybridFlow, a variational motion estimation framework for large displacements and deformations. A multi-scale hybrid matching approach is performed on the image pairs. Coarse-scale clusters formed by classifying pixels according to their feature descriptors are matched using the clusters' context descriptors. We apply a multi-scale graph matching on the finer-scale superpixels contained within each matched pair of coarse-scale clusters. Small clusters that cannot be further subdivided are matched using localized feature matching. Together, these initial matches form the flow, which is propagated by an edge-preserving interpolation and variational refinement. Our approach does not require training and is robust to substantial displacements and rigid and non-rigid transformations due to motion in the scene, making it ideal for large-scale imagery such as Wide-Area Motion Imagery (WAMI). More notably, HybridFlow works on directed graphs of arbitrary topology representing perceptual groups, which improves motion estimation in the presence of significant deformations. We demonstrate HybridFlow's superior performance to state-of-the-art variational techniques on two benchmark datasets and report comparable results with state-of-the-art deep-learning-based techniques.

\end{abstract}
\begin{document}

\flushbottom
\maketitle
%
%
\thispagestyle{empty}

\begin{figure}[!ht]
\centering
    \begin{subfigure}[t]{0.23\textwidth}
		\centering	
		\includegraphics[width=\textwidth]{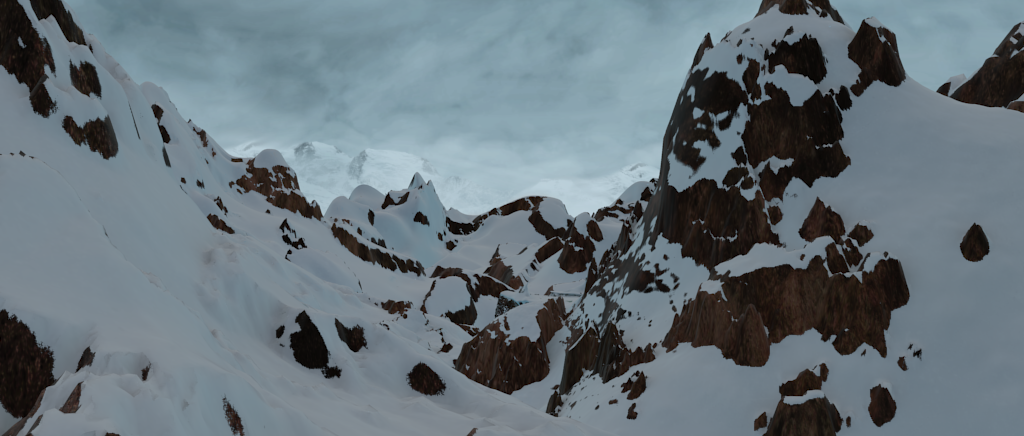}
		\caption{}
		\label{fig:mountain_input_frame}
	\end{subfigure}
	\begin{subfigure}[t]{0.23\textwidth}
		\centering	
		\includegraphics[width=\textwidth]{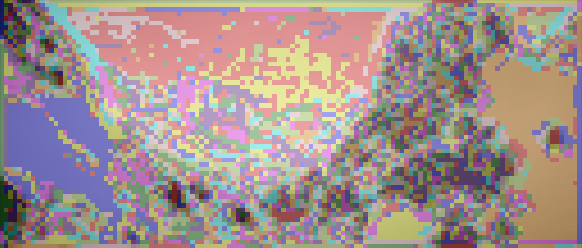}
		\caption{}
		\label{fig:mountain_labelmap}
	\end{subfigure}
	\begin{subfigure}[t]{0.23\textwidth}
		\includegraphics[width=\textwidth]{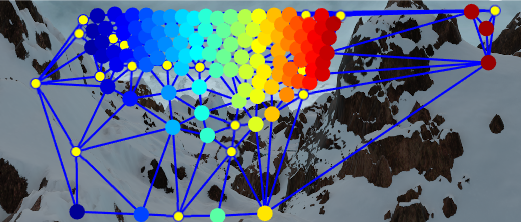}
		\caption{}
	    \label{fig:mountain_graph_matching_in_cluster(Left)}
	\end{subfigure}
	\begin{subfigure}[t]{0.23\textwidth}
		\includegraphics[width=\textwidth]{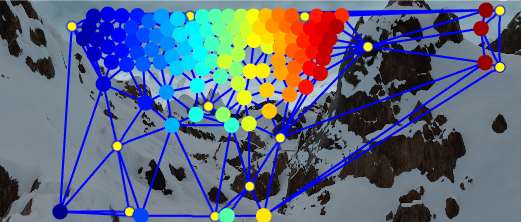}
		\caption{}
	    \label{fig:mountain_graph_matching_in_cluster(Right)}
	\end{subfigure}
	
	\begin{subfigure}[t]{0.23\textwidth}
		\includegraphics[width=\textwidth]{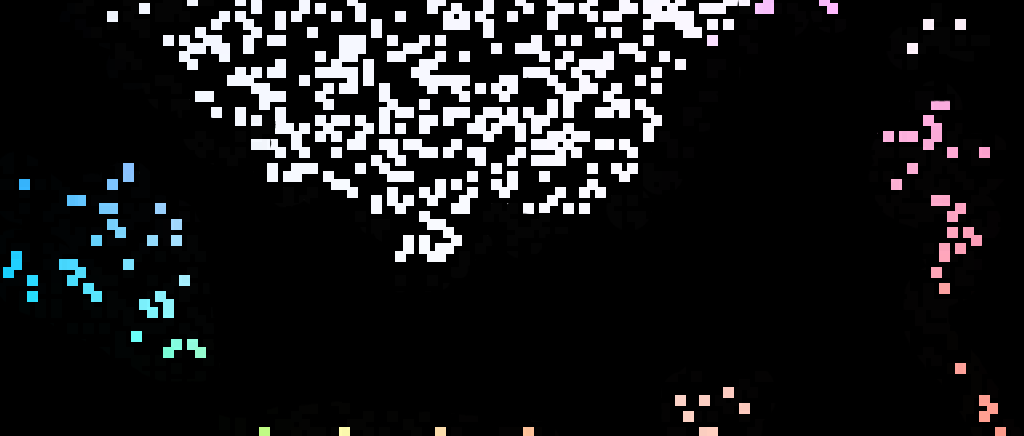}
		\caption{}
		\label{fig:mountain_initial_graphMatching_flow}
	\end{subfigure}
	\begin{subfigure}[t]{0.23\textwidth}
		\includegraphics[width=\textwidth]{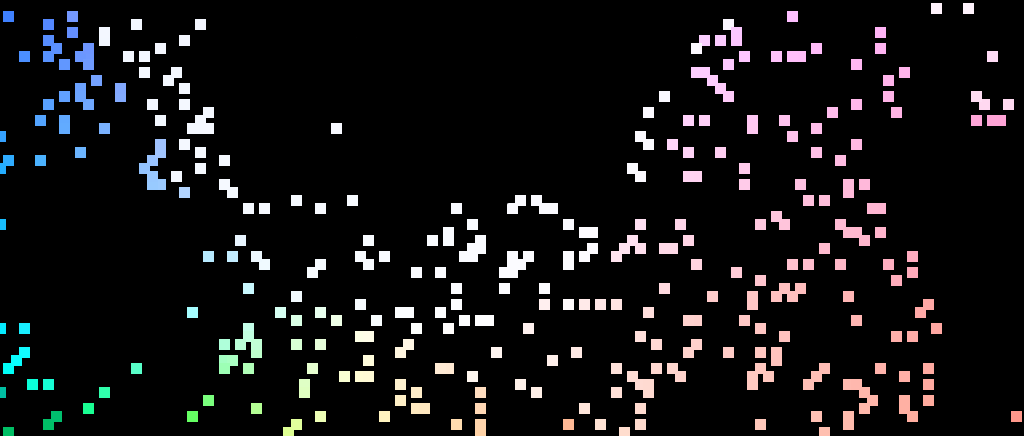}
		\caption{}
		\label{fig:mountain_initial_samllClusters_flow}
	\end{subfigure}
	\begin{subfigure}[t]{0.23\textwidth}
		\centering	
		\includegraphics[width=\textwidth]{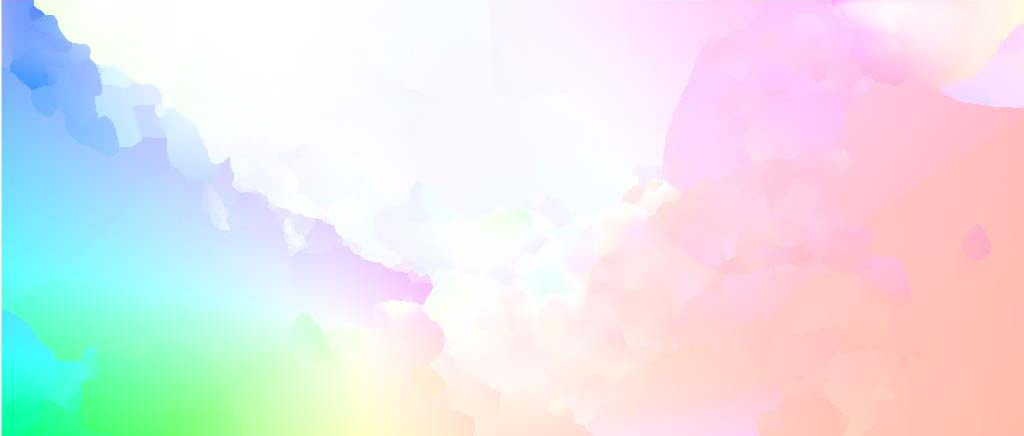}
		\caption{}
		\label{fig:mountain_interpolated_flow}
	\end{subfigure}
	\begin{subfigure}[t]{0.23\textwidth}
		\centering	
		\includegraphics[width=\textwidth]{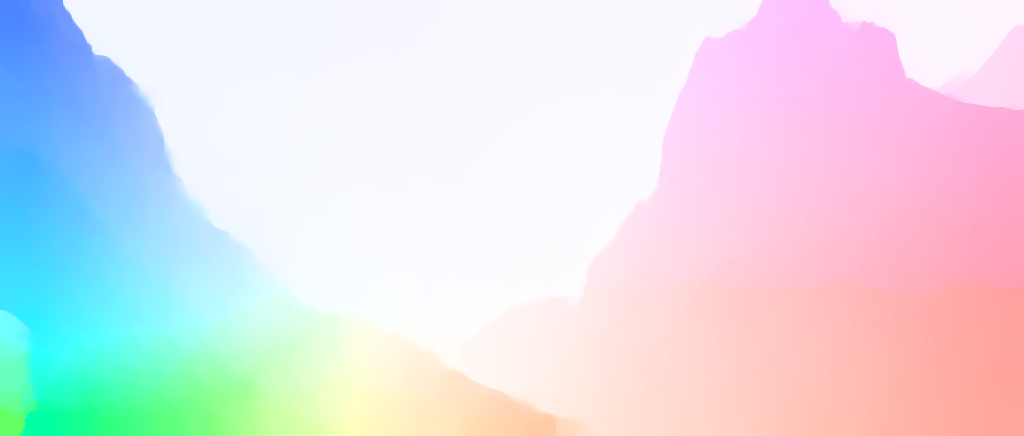}
		\caption{}
		\label{fig:mountain_refined_flow}
	\end{subfigure}
	\caption{
	(a) Input image frame. (b) Coarse-scale clusters from pixels' feature descriptors. (c) Color-coded graph matches of one coarse-scale cluster; first frame (a). (d) Color-coded graph matches for (c); second frame. (e) Motion vectors from graph-matching of superpixels at finest-scale. (f) Motion vectors from pixel feature matching in small clusters. (g) Interpolated flow from the combined initial motion vectors (e) + (f). (h) Final optical flow after variational refinement. Average Endpoint Error (EPE) = 0.157. Note: The pixels in (c) and (d) are magnified by $10\times10$ for clarity in the visualization.}
	\label{fig:main_page_mountainFig}
\end{figure}
\setlength{\textfloatsep}{10pt plus 1.0pt minus 2.0pt}

\begin{figure}[!ht]
\centering
\includegraphics[width=\textwidth]{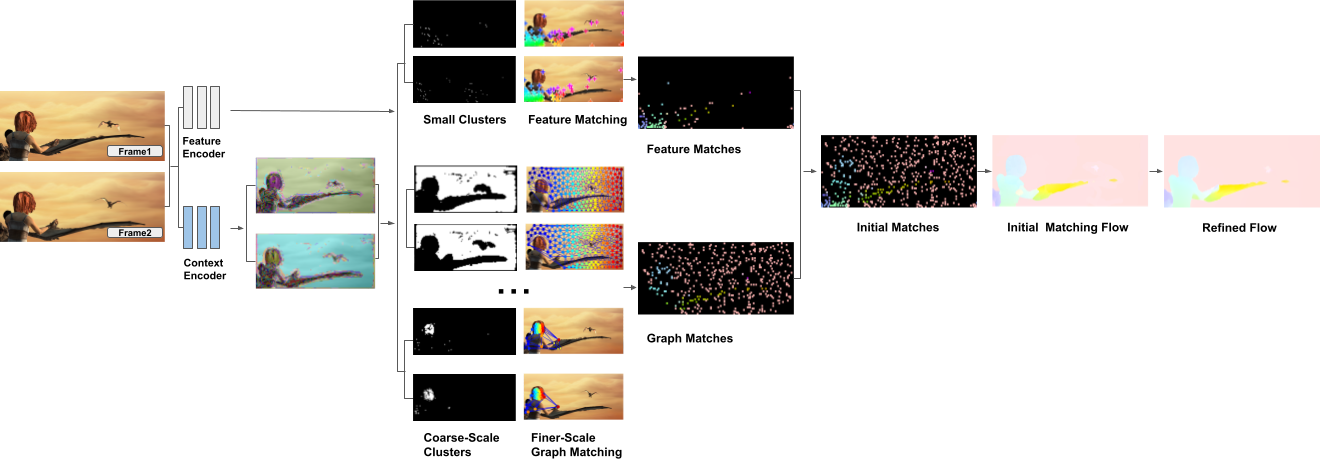}
\caption{HybridFlow: A multi-scale hybrid matching approach is performed on the image pairs. Uniquely, HybridFlow, leverages the strong discriminative nature of feature descriptors, combined with the robustness of graph matching on arbitrary graph topologies. Coarse-scale clusters are formed based on the pixels' feature descriptors and are further subdivided into finer-scale SLIC superpixels. Graph matching is performed on the superpixels contained within the matched coarse-scale clusters. Small clusters that cannot be further subdivided are matched using localized feature matching. Together, these initial matches form the flow, which is propagated by an edge-preserving interpolation and variational refinement.}
\label{fig:system_overview}
\end{figure}

\section{Introduction}
Dense motion estimation from optical flow is an essential component in many diverse computer vision applications ranging from autonomous driving \cite{wang2019unos}, multi-object tracking and segmentation \cite{porzi2020learning}, action recognition \cite{piergiovanni2019representation}, to video stabilization \cite{yu2020learning}, to name a few. Consequently, optical flow estimation directly contributes to the performance and accuracy of these applications.

Research in dense motion estimation techniques has been ongoing since the 1950s when Gibson first proposed it in \cite{gibson1950perception}. Despite the active research, to this day, the estimation of optical flow remains an open research problem. This is primarily attributed to the following two challenges: occlusions and large displacement. 

Occlusions can appear in several forms; self-occlusion, inter-object occlusion, or background occlusion. Typical solutions based on a variational approach employ a robust penalty function, and regularizers that aim to reduce the occlusion errors \cite{hur2019iterative, luo2019every}. However, they still fail in cases where the pixels vanish between consecutive frames. More recently, many deep-learning-based techniques were proposed \cite{liu2019selflow, bar2020scopeflow}. In many cases where ground truth is available, their performance surpasses that of variational techniques on benchmark datasets; however, applying these networks on real image sequences is a non-trivial task that requires re-training, fine-tuning and often manual annotation.

On the other hand, for large displacements, solutions follow a coarse-to-fine model that introduces additional errors due to the coarse scales' upsampling and interpolation. To alleviate some of the interpolation errors, Revaud et al. \cite{revaud2015epicflow} proposed EpicFlow, an edge-preserving interpolation of sparse matches used to initialize the optical flow motion estimation in a variational approach. Several techniques employing EpicFlow have since been proposed \cite{hu2016efficient, hu2017robust}, which address the sensitivity to noise in the sparse matches. The result is reduced interpolation errors in the estimated optical flow at the cost of over-smoothing the fine structures and failure to capture small-scale and fast-moving objects in the image. Thus, the accuracy of the initial sparse matches has a detrimental effect on the accuracy of the optical flow.

This paper presents HybridFlow (Figure \ref{fig:system_overview}), a robust variational motion estimation framework for large displacements and deformations based on multi-scale hybrid matching. Uniquely, HybridFlow leverages the strong discriminative nature of feature descriptors, combined with the robustness of graph matching on arbitrary topologies. We classify pixels according to the argmax of their context descriptor and form coarse-scale clusters. We follow a multi-scale approach, and fine-scale superpixels resulting from the perceptual grouping of pixels contained within the parent coarse-scale cluster form the basis of subsequent processing. Graph matching is performed on the graphs representing the fine-scale superpixels by simultaneously estimating the graph node correspondences based on the first and second-order similarities and a smooth non-rigid transformation between nodes. Graph matching is an NP-hard problem; thus, the graphs' factorization into Kronecker products ensures tractable computational complexity. This process can be repeated at multiple scales to handle arbitrarily large images. At the finest-scale, the pixels' feature descriptors are matched based on their $\mathcal{L}_{2}$ distance. Pixel-level feature matching is also performed on clusters that are too small to be subdivided into superpixels. We combine both sets of pixel matches to form the initial sparse motion vectors from which the optical flow is interpolated. Finally, variational refinement is applied to the optical flow. HybridFlow is robust to large displacements and deformations and has a minimal computational footprint compared to deep-learning-based approaches. A significant advantage of our technique is that using multi-scale graph matching reduces the computational complexity from $\mathcal{O}(n^{2})$ to $\sum_{i=0}^{k} \mathcal{O}(k^2)$ where $k$ is always smaller than the superpixel size $|s|$ and significantly smaller than $n$, i.e. $k < |s| << n$. Our experiments demonstrate the effectiveness of our technique in optical flow estimation. We evaluate HybridFlow on two benchmark datasets (MPI-Sintel \cite{butler2012naturalistic}, KITTI-2015\cite{kittiMenze2015ISA}) and compare it against state-of-the-art variational techniques. Hybridflow, outperforms all other variational techniques and, on average, gives comparable results with deep-learning-based methods.
 

\noindent
To summarize, our contributions are:
\begin{itemize}
    \setlength\itemsep{0.0em}
    \item A hybrid matching approach that uniquely combines the robustness of feature detection and matching with the invariance to rigid and non-rigid transformations of graph matching. The combination results in high tolerance to large displacements and deformations when compared to other techniques.
	\item An objective function based on first and second-order similarities for matching graph nodes and edges, which results in improved matching as showcased by our experiments.
	\item A complete variational framework for estimating optical flow that does not require training and is robust to large displacements and deformations caused due to motion in the scene while providing superior performance to state-of-the-art variational techniques and comparable performance to state-of-the-art deep-learning-based techniques on benchmark datasets.
\end{itemize}


\section{Related Work}
\label{sec:related_work}
Optical flow is a 2D vector field describing the apparent motion of the objects in the scene. This optical flow field can be very informative about the relations between the viewers' motion and the 3D scene.

Over the years, many techniques have been proposed following the predominant way of estimating optical flow using variational methods \cite{horn1981determining}. The optical flow is estimated via optimization of an energy model conditioned on image brightness/colour, gradient, and smoothness. This energy model fails when dealing with large displacements due to motion in the scene because its solution is approximate and locally optimizes the function. 

To address this challenge,  Anandan \cite{anandan1989computational} proposed a coarse-to-fine scheme. Coarse-to-fine techniques upsample and interpolate the flow from the finer-scale of the pyramid to the coarser. These techniques can deal with large displacement; however, it comes at the cost of over-smoothing any fine structures and failing to capture small-scale and fast-moving objects. 

At the same time, researchers explored the integration of feature matching in optical flow estimation. Revaud et al. \cite{revaud2016deepmatching} recently presented one of the most promising variational techniques where a HOG descriptor was used as a feature matching term in the energy function. Their technique can deal with deformations and is robust to repetitive textures. In subsequent work, the authors proposed EpicFlow, which performs a sparse-to-dense interpolation on the correspondences and estimates optical flow while preserving edges \cite{revaud2015epicflow}.  Hu et al. \cite{hu2017robust} built upon this work and proposed a robust interpolation technique to address the sensitivity of EpicFlow to noise in the initial matches by enforcing matching neighbourhood flow in the two images and fitting an affine model to the sparse correspondences. Up to now, this improvement produced superior performance than the previous best, which was based on a coarse-to-fine technique using PatchMatch \cite{hu2016efficient}.

More recently, several techniques were proposed based on convolutional neural networks (CNN). These estimate the optical flow in an end-to-end fashion using supervised learning \cite{flowNet2, PWCnet, SpyNet} or unsupervised learning \cite{ren2017unsupervised, UnFlow, geoNet}. One of the recent top-performing CNN-based approaches is SelFlow \cite{selflow}. SelFlow is a self-supervised learning approach for optical flow that, until lately, produced the highest accuracy among all unsupervised learning methods. The authors achieved this by creating synthetic occlusions from perturbing superpixels. The current state-of-the-art CNN-based technique is RAFT\cite{raft2021-662}, in which per-pixel features are employed in a deep network architecture of recurrent transforms. RAFT and its variants such as GMA\cite{GMA2021} currently achieve the best performance reporting the lowest average endpoint error for all significant optical flow benchmark datasets. 

Currently, the average endpoint error (AEE/EPE) reported on Sintel-final for the top-performing deep-learning technique (CRAFT) is 2.424, and for the top-performing variational technique (Hybridflow-ours) is 5.121; a difference of fewer than 2.7 pixels over the entire imageset of 562 images of 1024x436. Although deep learning techniques beget superior performance to the variational methods on benchmark datasets for which ground truth is available, they are unusable on real image sequences that seldom have associated ground truth, and training and fine-tuning become impossible. Moreover, even in cases where ground-truth may be available, the training and fine-tuning are time-consuming, offline operations that render them unsuitable in scenarios requiring real or interactive time performance. 

For these reasons, we propose a variational optical flow technique that is independent of the content of the image sequences and does not impose additional requirements for training and fine-tuning. Our method follows a hybrid approach for matching to eliminate errors in the initial sparse matches introduced from large displacements and deformations. HybridFlow leverages the strong discriminative nature of feature descriptors combined with the robustness of deformable graph matching. In contrast to variational state-of-the-art, which employs a regular grid structure in their coarse-to-fine matching scheme, HybridFlow operates at only a single image scale and multiple scales of clustering, eliminating over-smoothing and handling small-scale and fast-moving objects better. More notably, our method does not restrict deformations by enforcing smooth neighbourhood matching but instead employs deformable graph matching, which allows for rigid and non-rigid transformations between neighbouring superpixels. 

\section{Graph Model and Matching}
\label{sec:graph_model}
\textbf{Model.} A graph $G = \{P, E, T\}$ consists of nodes $P$ inter-connected with edges $E$. A node-edge incidence matrix $T$ specifies the topology of the graph $G$. The nodes are represented in matrix form as $ P  = \left[ \vec{p_{1}}, \vec{p_{2}}, \dots, \vec{p_{N}} \right[ \in \mathbb{R}^{dim(\vec{p}) \times N}$, where $dim: \vec{v} \longrightarrow \mathbb{R}$ is a function that returns the cardinality of a vector $\vec{v}$. Similarly, the edges are represented in matrix form as $ E = \left[ \vec{e_{1}}, \vec{e_{2}}, \dots, \vec{e_{M}} \right[ \in \mathbb{R}^{dim(\vec{e}) \times M}$. An edge-weight function $w: E \times E \longrightarrow \mathbb{R}$ assigns weights to edges. Given the above definitions, the incidence matrix is defined as $T \in \{0,1\}^{N\times M}$ where $T_{(i,k)} = T_{(j,k)} = 1$, if an edge $e_{k} \in E$ connects the nodes $p_{i}, p_{j} \in P$, otherwise it is set to 0.

\begin{minipage}{0.45\textwidth}
    \centering
	\includegraphics[width=\textwidth]{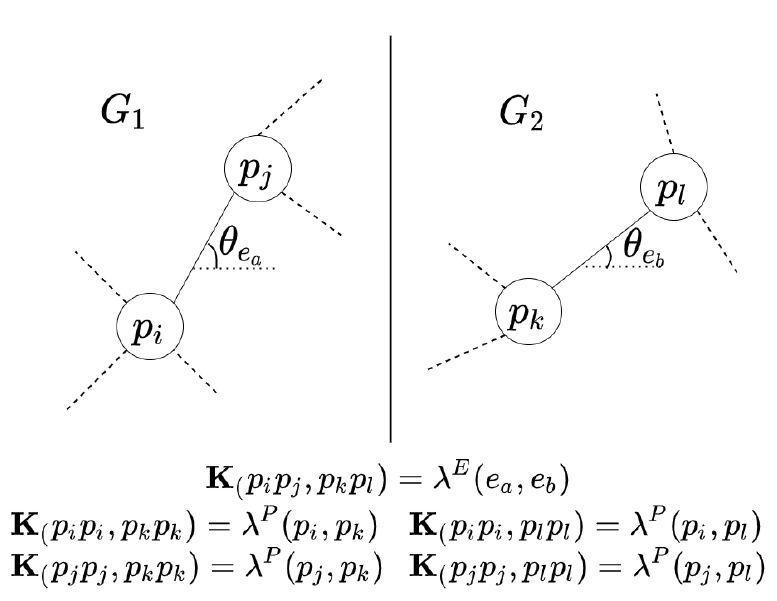}
	\textbf{Figure 3}. Two nodes in $G_{1}$ and $G_{2}$. The element values in $\textbf{K}$ are calculated according to Equations \ref{eq:similarity_measure_P} and \ref{eq:similarity_measure_E}.
	\label{fig:graph_example}
\end{minipage}
\begin{minipage}{0.55\textwidth}
\textbf{Matching.} Matching two graphs $G_{1} = \{P_{1}, E_{1}, T_{1}\}$ and $G_{2} = \{P_{2}, E_{2}, T_{2}\}$ is an NP-hard problem for which exact solutions can only be found if the number of nodes and edges are significantly small e.g. $N, M < 15$. Proposed solutions typically formulate graph matching as a Quadratic Assignment Problem(QAP) and provide an approximation to the solution \cite{dokeroglu2019artificial}. This requires the calculation of two affinity matrices: $A^{P}_{1,2} \in \mathbb{R}^{N\times N}$ which encodes the similarities between nodes in $G_{1}$ and $G_{2}$, and $A^{E}_{1,2} \mathbb{R}^{M\times M}$ which encodes the similarities between edges in $G_{1}$ and $G_{2}$. The functions $\lambda^{P}: P \times P \longrightarrow \mathbb{R}$ and $\lambda^{E}: E \times E \longrightarrow \mathbb{R}$ measure the similarities between nodes and edges, respectively. Therefore for two corresponding nodes $p_{i} \in P_{1}$ of $G_{1}$ and $p_{k} \in P_{2}$ of $G_{2}$, the node affinity matrix element is $A^{P}_{i,k} = \lambda^{P}(p_{i}, p_{k})$. Similarly, for edges $e_{a} \in E_{1}$ of $G_{1}$ and $e_{b} \in E_{2}$ of $G2$ the edge affinity matrix element is $A^{E}_{a,b} = \lambda^{E}(e_{a}, e_{b})$.        
\end{minipage}

\vspace{10pt}
Given the above definitions, the solution to matching $G_{1}$ and $G_{2}$ is equivalent to finding the correspondence matrix $C_{1,2} \in \{0,1\}^{N_{1}\times N_{2}}$ between the nodes of $G_{1}$ and $G_{2}$, that maximizes,
\begin{equation}
	\argmax_{C_{1,2} \in \{0,1\}^{N_{1}\times N_{2}}} \textbf{1}_{C_{1,2}}^T\textbf{K}\textbf{1}_{C_{1,2}}
	\label{eq:optimization}
\end{equation}
where $\textbf{1}_{C_{1,2}} \in \{0,1\}^{N_{1}\times N_{2}}$ is the characteristic function, and $\textbf{K} \in \mathbb{R}^{N_{1}N_{2}\times N_{1}N_{2}}$ is a composite affinity matrix that combines the node affinity matrix $A^{P}_{1,2}$ and the edge affinity matrix $A^{E}_{1,2}$. The element of $\textbf{K}((p_{i}p_{j})_{1}, (p_{k}p_{l})_{2})$ for the nodes $p_{i}, p_{j} \in P_{1}$, $p_{k}, p_{l} \in P_{2}$, and the edges connecting these nodes $e_{a} \in E_{1}$, $e_{b} \in E_{2}$ respectively, is calculated as,
\begin{equation}
K((p_{i}p_{j})_{1}, (p_{k}p_{l})_{2}) =
\begin{cases}
      \lambda^{P}(p_{i}, p_{k}) & \text{if } p_{i} = p_{j} \text{ and } p_{k} = p_{l},\\
      \lambda^{E}(e_{a}, e_{b}) & \text{if } p_{i} \neq p_{j} \text{ and } p_{k} \neq p_{l},\\
      0 & \text{otherwise}
    \end{cases}     
    \label{eq:element_K}
\end{equation}

An example is shown in Figure \ref{fig:graph_example}. Intuitively, if the two nodes considered in each graph are co-located, i.e. there is \textit{no edge} connecting them, then the element's value is the similarity of the function $\lambda^{P}(.,.)$ for the nodes. If the two nodes are different, i.e. there \textit{is} an edge connecting them, then the element's value is the similarity of the function $\lambda^{E}(.,.)$ for the connecting edges; otherwise, it is set to 0. 

\section{Method}
\label{sec:method}
Figure \ref{fig:system_overview} and Algorithm \ref{alg:pipeline} summarize the steps of the proposed technique. HybridFlow is the refined flow resulting from the interpolation of the combined initial flows calculated from the sparse graph matches from superpixels and feature matches of pixels in small clusters, as explained below. 
\begin{algorithm}
	\SetAlgoLined
	\KwResult{Optical flow $\mathcal{O}$ between image-pair $I_{1}, I_{2}$}
		1. initialize optical flow $\mathcal{O} = \{ \}$\;
		
		2. pixel classification (Eq. \ref{eq:argmax_class}), clustering (Sec. \ref{subsec:feature_matching}), and matching of $\{\mathcal{C}_{1}^{1}, \ldots, \mathcal{C}_{n}^{1}\} \in I_{1}$, $\{\mathcal{C}_{1}^{2}, \ldots, \mathcal{C}_{n}^{2}\} \in I_{2}$ $ \rightarrow \mathcal{M}_{cluster}$\;
        
        3. \For{$(\mathcal{C}_{i}^{1}, \mathcal{C}_{i}^{2})$ in $\mathcal{M}_{cluster}$}		{
        \uIf {$|\mathcal{C}_{i}^{1}| > 10,000$ and $|\mathcal{C}_{i}^{2}| > 10,000$}
        {a. $(\mathcal{C}_{i}^{1}, \mathcal{C}_{i}^{2})$ $\rightarrow \mathcal{M}_{coarse}$\;
         b. fine-scale clustering with SLIC $\rightarrow  \mathcal{S}_{i}^{1} \subset \mathcal{C}_{i}^{1} \in I_{1}, \mathcal{S}_{i}^{2} \subset \mathcal{C}_{i}^{2} \in I_{2}$ (Sec. \ref{subsec:feature_matching})\;
    	 c. graph matching of superpixels in $\mathcal{S}_{i}^{1}, \mathcal{S}_{i}^{2} \rightarrow \mathcal{M}_{fine}$ (Sec. \ref{subsec:graph_matching})\;
		 d. sparse flow $\mathcal{O}_{c}$ from pixel matching within each matched pair $(\mathcal{S}_{i}^{1}, \mathcal{S}_{i}^{2}) \in \mathcal{M}_{fine}$\;
		 
        } 
        \uElse{$(\mathcal{C}_{i}^{1}, \mathcal{C}_{i}^{2})$ $\rightarrow \mathcal{M}_{small}$\;
        a. sparse flow $\mathcal{O}_{s}$ from pixel matching within matched pair $(\mathcal{C}_{i}^{1}, \mathcal{C}_{i}^{2}) \in \mathcal{M}_{small}$\;} 
        }
        4. initial sparse flow $\mathcal{O} = \mathcal{O} \cup \{\mathcal{O}_{c},  \mathcal{O}_{s}\} $\;
		5. interpolation of sparse flow $\mathcal{O}$ and variational refinement(Sec. \ref{subsec:interpolation_and_refinement})\;
 \caption{HybridFlow}
 \label{alg:pipeline}
\end{algorithm}
\setlength{\textfloatsep}{10pt plus 1.0pt minus 2.0pt}
\noindent


\subsection{Perceptual Grouping and Feature Matching}
\label{subsec:feature_matching}
Feature descriptors encode discriminative information about a pixel and form the basis of the perceptual grouping and matching. We conduct experiments with three different feature descriptors: rootSIFT proposed in \cite{ArandjelovicZisserman:2012}, pretrained DeepLab on ImageNet, and pretrained encoders with the same architecture as in \cite{raft2021-662}. 
As discussed later in the experimental results and Section \ref{subsec:implementation}, the latter descriptor results in the best performance. Next, we cluster pixels based on their feature descriptors to replace the rigid structure of the pixel grid as shown in Figure \ref{fig:mountain_labelmap}. Specifically, we classify each pixel as the argmax value of its N-dimensional feature descriptor and aggregate them into clusters. Thus, a pixel $p$ is assigned a cluster index $i_{p}$ given by,
\begin{equation}
i_{p} = \argmax(Sigmoid(Softmax(ReLU({F}_{c}(p)))))
\label{eq:argmax_class}
\end{equation}
where $\mathcal{F}_{c}$ is the feature descriptor. 
Hence, this results in a set of coarse-scale clusters in each image matched according to their cluster indices. 

Pixels contained in clusters with an area less than $10,000$ are matched according to their feature descriptors. Outliers in the initial matches are removed from subsequent processing using RANSAC, which finds a localized fundamental matrix per cluster.


The initial sparse flow resulting from this step consists of the flow calculated from each of the inlier features. Figure \ref{fig:mountain_initial_samllClusters_flow} shows the initial flow resulting from the sparse feature matching of the pixels contained within all small clusters. The size of pixels is magnified by $10\times10$ for clarity in the visualization. 

Coarse-scale clusters with a larger area than $10,000$ pixels are further clustered by a simple linear iterative clustering (SLIC) which adapts k-means clustering to group pixels into perceptually meaningful atomic regions \cite{achanta2012slic}. The parameter $\kappa$ is calculated based on the image size and the desired superpixel size and is given by $\kappa = \frac{|I|}{|s|}$ where $|s| \approx 2223, s \in \mathcal{S}$, and $|I|$ is the size of the image. This restricts the number of the approximately equally-sized superpixels $\mathcal{S}$; in our experiments discussed in Section \ref{subsec:implementation}, the optimal value for $\kappa$ $\approx 250$ to $300$. For the finer-scale superpixels $\mathcal{S}$, a graph is constructed where each node corresponds to a superpixel's centroid, and edges correspond to the result Delaunay triangulation as explained in the following Section \ref{subsec:graph_matching}.

\subsection{Graph Matching}
\label{subsec:graph_matching}
The two sets of superpixels contained in the matched coarse-scale clusters  of images $I_{1}, I_{2}$ are represented with the graph model described in Section \ref{sec:graph_model}. For each superpixel $S$, the nodes $P$ are a subset of all the pixels $p$ in $S$ i.e. $P \subseteq \{p : \forall p \in S \in I\}$. The edges $E$ and topology $T$ of each graph are derived from a Delaunay triangulation of the nodes $P$. The graph is undirected, and the edge-weight function $w(.,.)$ is symmetrical w.r.t. edges $\vec{e_{a}}, \vec{e_{b}} \in E$, such that $w(\vec{e_{a}}, \vec{e_{b}}) = w(\vec{e_{b}}, \vec{e_{a}})$. The similarity functions $\lambda^{P}(.,.)$ and $\lambda^{E}(.,.)$ are also symmetrical; for $p_{i}, p_{j} \in P_{1}$, $p_{k}, p_{l} \in P_{2}$, and edges $e_{a} \in E_{1}$, $e_{b} \in E_{2}$, the similarity functions are given by,
\begin{equation}
	\lambda^{P}	(p_{i}, p_{k}) = e^{-|d^{P}(f(p_{i}), f(p_{k}))|} \\
	\label{eq:similarity_measure_P}
\end{equation}
\begin{equation}
	\lambda^{E}	(e_{a}, e_{b}) = e^{ - \frac{1}{2}\left[ \Phi^{\circ} + 
	|d^{E}(\theta_{e_{a}}, \theta_{e_{b}})| + |d^{L}(e_{a}, e_{b})|  \right]} 
	\label{eq:similarity_measure_E}
\end{equation}
where $\Phi^{\circ}$ is given by,
\begin{equation}
    \begin{aligned}
    \Phi^{\circ} = \Phi^{1}_{gradient}(f(p_{i}), f(p_{j}), f(p_{k}), f(p_{l})) +	
	\Phi^{2}_{gradient}(f(p_{i}), f(p_{j}), f(p_{k}), f(p_{l})) +	\\
	\Phi^{1}_{color}(\mathcal_{C}(p_{i}), \mathcal_{C}(p_{j}), \mathcal_{C}(p_{k}), \mathcal_{C}(p_{l})) +	
	\Phi^{2}_{color}(\mathcal_{C}(p_{i}), \mathcal_{C}(p_{j}), \mathcal_{C}(p_{k}), \mathcal_{C}(p_{l}))
	\label{eq:combined}
	\end{aligned}
\end{equation}
\begin{equation}
    \begin{aligned}
	\Phi^{1}_{gradient}= | d^{P}(f(p_{i}), f(p_{k})) | +  | d^{P}(f(p_{j}), f(p_{l})) |
	\\
	\Phi^{1}_{color}= | d^{\mathcal{C}}(f(p_{i}), f(p_{k})) | +  | d^{\mathcal{C}}(f(p_{j}), f(p_{l})) |
	\label{eq:first_order}
	\end{aligned}
\end{equation}

\begin{equation}
    \begin{aligned}
    \Phi^{2}_{gradient}= | d^{P}(f(p_{i}), f(p_{j})) | -  | d^{P}(f(p_{k}), f(p_{l})) |
	\\
	\Phi^{2}_{color}= | d^{\mathcal{C}}(f(p_{i}), f(p_{j})) | -  | d^{\mathcal{C}}(f(p_{k}), f(p_{l})) |
	\label{eq:second_order}
	\end{aligned}
\end{equation}
$f: P \longrightarrow S$ is a feature descriptor with cardinality $S$ for a node $p \in P$, $\mathcal{C}: P \longrightarrow 6$ is a function which calculates the 6-vector $<\mu_{r}, \mu_{g}, \mu_{b}, \sigma_{r}, \sigma_{g}, \sigma_{b}>$ containing color distribution means and variances ($\mu, \sigma$) at $p$ modeled as a 1D Gaussian for each color channel, $d^{P}: S \times S \longrightarrow \mathbb{R}$ is the $\mathcal{L}^{1}$-norm of the difference between the feature descriptors of two nodes in $p_{i}, p_{j}, p_{k}, p_{l} \in P$, $d^{E}: \mathbb{R} \times \mathbb{R} \longrightarrow \mathbb{R}$ is the difference between the angles $\theta_{e_{a}}, \theta_{e_{b}}$ of the two edges $e_{a}\in E_{1}, e_{b}\in E_{2}$ to the horizontal axes, and $d^{\mathcal{C}}: 6 \times 6 \longrightarrow \mathbb{R}$ is the $\mathcal{L}^{1}$-norm of the difference between the two 6-vectors containing color distribution information for the two nodes in $p_{i}, p_{j}, p_{k}, p_{l} \in P$. 

$\Phi^{1}_{*}$ signify first-order similarities and measures similarities between the nodes and edges of the two graphs. In addition to the first-order similarities $\Phi^{1}_{*}$, the functions in the above equations define additional second-order similarities $\Phi^{2}_{*}$ which have been shown to improve the performance of the matching \cite{cho2010reweighted}. That is, instead of using only similarity functions that result in small differences between similar gradients/colours and large otherwise, e.g. first-order, we additionally incorporate the second-order similarities defined above, which measure the similarity between the two gradients and colours using the \textit{distance between their differences} \cite{tian2019sosnet}. For example, the first-order similarity $\Phi^{1}_{gradient}$ calculates the distance between the two feature descriptors in the two graphs i.e. $\lambda^{P}(p_{i}, p_{k})$ in Equation \ref{eq:similarity_measure_P}, whereas the second-order similarity calculates the \textit{distance between the feature descriptor differences of the end-points in each graph} i.e. $\Phi^{2}_{gradient}$ and $\Phi^{2}_{color}$ in Equations \ref{eq:similarity_measure_P} and \ref{eq:second_order}. A descriptor $f(s_{i})$, as defined in Equation \ref{eq:combined}, is calculated for each centroid-node representing superpixel $s_{i} \in \mathcal{S}$ as the average of the feature descriptors of all pixels contained within it $f(s_{i}) = \frac{1}{|s_{i}|} \sum_{\forall p\in s_{i} \subset I} \phi_{p}$ where $|s_{i}|$ is the number of pixels in superpixel $s_{i}$, and $\phi_{p}$ is the feature descriptor of pixel $p\in s_{i} \subset I$. 

Given the above function definitions, graph matching is solved by maximizing Equation \ref{eq:optimization} using a path-following algorithm. $\textbf{K}$ is factorized into a Kronecker product of six smaller matrices which ensures tractable computational complexity on graphs with nodes $N, M \approx 300$ \cite{zhou2012factorized}. Furthermore, robustness to geometric transformations such as rotation and scale is increased by finding an optimal transformation at the same time as finding the optimal correspondences and thus enforcing global rigid (e.g. similarity, affine) and non-rigid geometric constraints during the optimization \cite{zhou2013deformable}. 

The result is superpixels matches within the matched coarse-scale clusters. Assuming a piecewise rigid motion, we use RANSAC to remove outliers from the superpixel matches. For each superpixel $s$ having at least three matched neighbours, we fit an affine transformation. We only check whether the superpixel $s$ is an outlier, in which case it is removed from further processing. This process is repeated for all small clusters and graph-matched superpixels. We proceed by matching the pixels contained within the matched superpixels based on their feature descriptors. Similar to earlier in Section \ref{subsec:feature_matching}, we remove outlier pixel matches contained in the superpixels using RANSAC to find a localized fundamental matrix. 

The initial sparse flow resulting from graph matching consists of flow calculated from every pixel contained in the matched superpixels. Figure \ref{fig:mountain_labelmap} shows the result of the clustering of the feature descriptors for the image shown in Figure \ref{fig:mountain_input_frame}. Clusters having a large area are further divided into superpixels. The graph nodes correspond to each superpixel's centroid, and the edges result from the Delaunay triangulation of the nodes, as explained above. Figure \ref{fig:mountain_graph_matching_in_cluster(Left)} and Figure \ref{fig:mountain_graph_matching_in_cluster(Right)} show the result of graph matching superpixels within a matched coarse-scale clusters. The matches are colour-coded, and unmatched nodes are depicted as smaller yellow circles.  Examples of unmatched nodes appear in the left part of the left image in Figure \ref{fig:mountain_graph_matching_in_cluster(Left)}. The images shown are from the benchmark dataset MPI-Sintel \cite{butler2012naturalistic}.

\subsection{Interpolation and Refinement}
\label{subsec:interpolation_and_refinement}
The combined initial sparse flows (Figure \ref{fig:mountain_initial_samllClusters_flow}, \ref{fig:mountain_initial_graphMatching_flow}) calculated from sparse feature matching and graph matching, as described above in Sections \ref{subsec:feature_matching} and \ref{subsec:graph_matching} respectively, are  first interpolated and then refined. For the interpolation, we apply an edge-preserving technique \cite{revaud2015epicflow}. This results in dense flow as shown in Figure \ref{fig:mountain_interpolated_flow}. In the final step, we refine the interpolated flow using variational optimization on the full-scale of the initial flows, i.e. no coarse-to-fine scheme, with the same data and smoothness terms as used in \cite{revaud2015epicflow}. The final result is shown in Figure \ref{fig:mountain_refined_flow}. 

\section{Experimental Results}
\label{sec:experimental_results}
In this section, we report on the evaluation of HybridFlow on benchmark datasets and compare it with state-of-the-art variational optical flow techniques. In the supplemental material, we present two applications of the proposed technique on large-scale image-based reconstruction where ground truth is unavailable. Specifically, we use Wide-Area Motion Imagery (WAMI), and Full-Motion Video (FMV) captured from aerial sensors and demonstrate how our technique easily scales to ultra-high resolution images, in contrast to deep learning alternatives.

\subsection{Datasets and evaluation metrics}
We evaluate HybridFlow on the two widely used benchmark datasets for motion estimation:
\begin{itemize}
    \setlength\itemsep{0em}
    \item MPI-Sintel\cite{butler2012naturalistic} --- a synthetic data set for the evaluation of optical flow derived from the open source 3D animated short film, Sintel. It includes image sequences with large displacements, motion blur, and non-rigid motion.
    \item KITTI-2015\cite{kittiMenze2015ISA} --- a real data set captured with an autonomous driving platform. It contains dynamic scenes of real world conditions and features large displacements and complex 3D objects.
\end{itemize}
The quantitative evaluation is performed in terms of the average endpoint error(EPE) for MPI-Sintel, and percentage of optical flow outliers(FI) for KITTI-2015.

\subsection{Implementation details}
\label{subsec:implementation}
\begin{figure}[ht!]
\centering
    \begin{subfigure}{0.32\textwidth}
    \includegraphics[width=\textwidth]{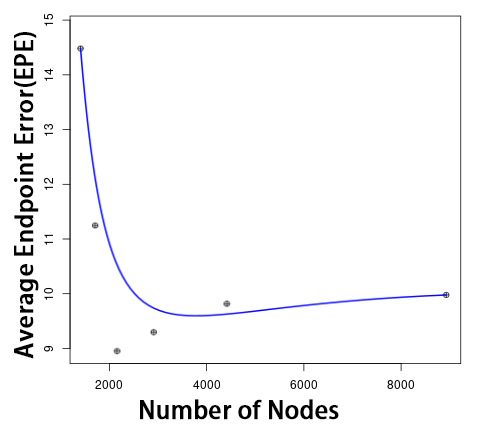}
    \caption{}
    \label{fig:graphMatchingEPE}
    \end{subfigure}
    \begin{subfigure}{0.32\textwidth}
    \includegraphics[width=\textwidth]{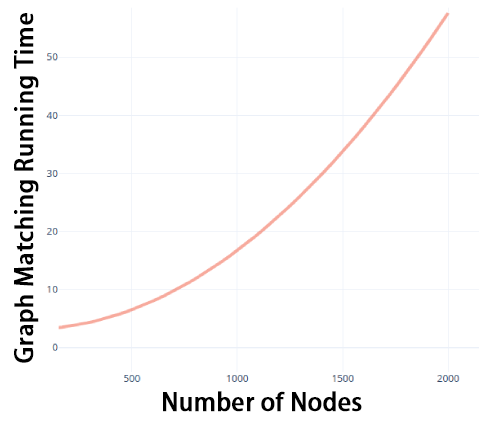}
    \caption{}
    \label{fig:graphMatchingAnalysis}
    \end{subfigure}
    \begin{subfigure}{0.32\textwidth}
    \includegraphics[width=\textwidth]{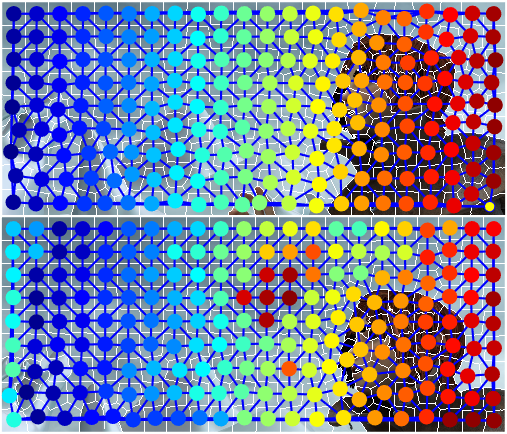}
    \caption{}
    \label{fig:graph_matching_occlussion_example}
    \end{subfigure} 
\caption{(a) Average end-point error (EPE) w.r.t number of graph nodes per image($1024 \times 436$).(b) Average graph-matching time complexity(seconds) w.r.t. number of graph nodes. We empirically determine the optimal number of superpixels by performing graph matching using different superpixel sizes and calculate the EPE of the resulting optical flow. Optimal size is found to be $|s| \approx 300$. (c) Ablation: Graph matching using SLIC clusters as the initial coarse-scale-clusters instead of clustering the feature descriptors. Superpixel clustering results in a near-rigid pixel grid that, as can be seen, is not robust to occlusions. The number of superpixels is set to 200. The first and second columns show the colour-coded matches of the graph nodes using graph matching based on an initial coarse-scale clustering of superpixels (SLIC).}    
\end{figure}

\begin{figure}[!ht]
\centering
     \begin{subfigure}[t]{0.23\textwidth}
         \includegraphics[width=\textwidth]{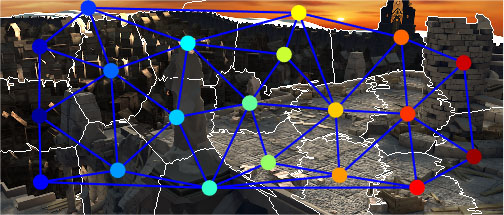}
         \label{fig:temple_20_fgm_left}
     \end{subfigure}
     \begin{subfigure}[t]{0.23\textwidth}
         \includegraphics[width=\textwidth]{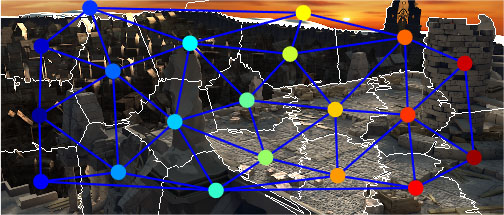}
         \label{fig:temple20fgmright}
     \end{subfigure}
     \begin{subfigure}[t]{0.23\textwidth}
         \includegraphics[width=\textwidth]{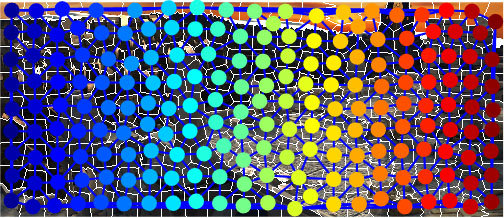}
         \label{fig:temple_200_fgm_left}
     \end{subfigure}
     \begin{subfigure}[t]{0.23\textwidth}
         \includegraphics[width=\textwidth]{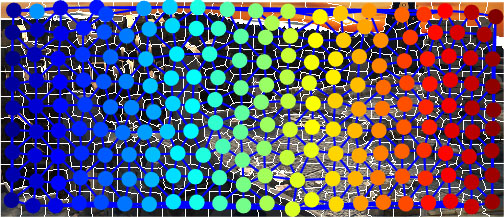}
         \label{fig:temple_200_fgm_right}
     \end{subfigure}
     
     \begin{subfigure}[t]{0.23\textwidth}
         \includegraphics[width=\textwidth]{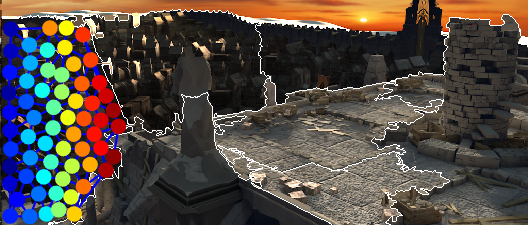}
         \label{fig:temple_200_fgm_left}
     \end{subfigure}
     \begin{subfigure}[t]{0.23\textwidth}
         \includegraphics[width=\textwidth]{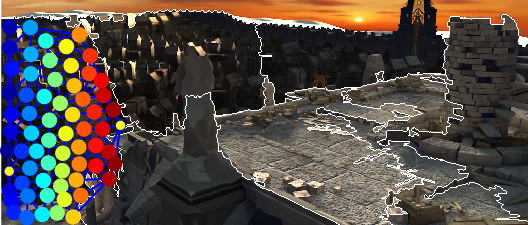}
         \label{fig:temple_200_fgm_right}
     \end{subfigure}
     \begin{subfigure}[t]{0.23\textwidth}
         \includegraphics[width=\textwidth]{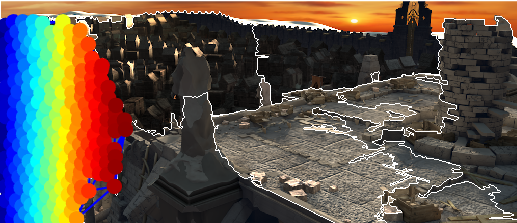}
         \label{fig:temple_200_fgm_left}
     \end{subfigure}
     \begin{subfigure}[t]{0.23\textwidth}
         \includegraphics[width=\textwidth]{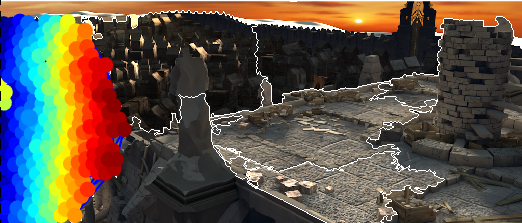}
         \label{fig:temple_200_fgm_right}
     \end{subfigure}
     \caption{Superpixel size. We performed graph-matching with different superpixel sizes. The rows correspond to the examples of superpixel sizes $|s| = 22323$(20 superpixels), $|s| = 2232$(200 superpixels), $|s| = 1116$(5 clusters subdivided into 80 superpixels) and $|s| = 223$(5 clusters subdivided into 400 superpixels)respectively. The first and second columns show colour-coded matches of the graph nodes using only graph matching as explained in Section \ref{subsec:graph_matching}.}
     \label{fig:superpixel_size}
\end{figure}

We have implemented the proposed approach in Python. All experiments were run on a workstation with an Intel i7 processor. We extract the features descriptors using the approach introduced in RAFT \cite{raft2021-662}. Perceptual grouping using SLIC superpixels is performed using the method in \cite{achanta2012slic}. We factorize graphs into Kronecker products as presented in \cite{zhou2012factorized} and perform deformable graph matching following the approach in \cite{zhou2013deformable}. Finally, we interpolate the combined initial flows from sparse feature matching and graph matching using the edge-preserving interpolation and variational refinement in EpicFlow\cite{revaud2015epicflow}.
\noindent
\textbf{Superpixel size.} We empirically determined the optimal size of the superpixels which subsequently determined the number of superpixels $\kappa$ as defined in Section \ref{subsec:feature_matching}. Figure \ref{fig:superpixel_size} shows an example from the experiments on different superpixel sizes. The rows correspond to the superpixel sizes $|s| = 22323$(20 superpixels), $|s| = 2232$(200 superpixels), $|s| = 1116$(400 superpixels) and $|s| = 223$(2000 superpixels) respectively. The first and second columns show the colour-coded matches using only the graph matching technique described in Section \ref{subsec:graph_matching}.  Figure \ref{fig:graphMatchingEPE} shows a graph of the average endpoint error (EPE) of the final optical flow as a function of the superpixel size performed on the training image sequences of the MPI-Sintel dataset. In Figure \ref{fig:graphMatchingAnalysis} we show the increase of the graph matching's computational time as a function of the number of nodes in the graphs. 

\setlength{\textfloatsep}{10pt plus 1.0pt minus 4.0pt}
\textbf{Initial coarse-scale clustering.}
The initial coarse-scale clusters are formed by clustering the pixels' feature descriptors. This is a crucial part of the process, which increases robustness to large displacements. As shown in Figure \ref{fig:graph_matching_occlussion_example}, using SLIC superpixels on the entire image results in a near-rigid rectangular pixel grid and consequently failures in graph matching. This is evident from the mismatching of the dark red circles in the middle of the right image. Our experiments show that an irregular pixel grid based on features descriptors increases the robustness in the presence of large displacements and deformations. 

\begin{table}[!htbp]
\centering
\caption{Benchmark datasets results. The top half of the table (\emph{DL}) are the top-performing deep learning methods; the bottom half of the table (\emph{VM}) are the top-performing variational methods. For MPI-Sintel results, EPE-noc is the EPE on the non-occluded areas, and EPE-occ is the EPE on occluded areas. s0-10 is the EPE for pixels whose motion speed is between 0-10 pixels, similarly for s10-40 and s40+. For the KITTI-2015 test-set non-occluded pixels, FI-bg is the percentage of optical flow outliers for background, FI-fg is the percentage of optical flow outliers for the foreground, FI-all/Est is the percentage of outliers averaged over all non-occluded ground truth pixels.}
\label{tab:table_results}
\resizebox{0.9\textwidth}{!}{
    \begin{tabular}{@{}llllllllllllllll@{}}
    \toprule
    & & \multicolumn{9}{l}{Sintel-final pass} & \multicolumn{2}{l}{Sintel-Clean} & \multicolumn{3}{l}{KITTI-20015} \\ \midrule
     \hspace{1cm} & Method & EPE all & EPE-noc & EPE-occ & d0-10 & d10-60 & d60-140 & s0-10 & s10-40 & s40+ & EPE all & EPE-noc & \begin{tabular}[c]{@{}l@{}}Fl-all\\ Noc/est \end{tabular} & \begin{tabular}[c]{@{}l@{}}Fl-fg\\ Noc/Est \end{tabular} & \begin{tabular}[c]{@{}l@{}}Fl-bg\\ Noc/Est\end{tabular} \\ \midrule     
    \multirow{1}{*}{\begin{tabular}[c]{@{}l@{}} \emph{DL \vspace{5mm}} \end{tabular}} 
	&  GMA\cite{GMA2021} & \textbf{2.470} & \textbf{1.241} & \textbf{12.501} & \textbf{2.863} & \textbf{1.057} & \textbf{0.653} & \textbf{0.566} & \textbf{1.817} & \textbf{13.492} & \textbf{1.388} & \textbf{0.582} & \textbf{2.94} & \textbf{3.69} & 3.07 \\ \cmidrule(l){2-16} 
    & RAFT\cite{raft2021-662} & 2.855 & 1.405 & 14.680 & 3.112 & 1.133 & 0.770 & 0.634 & 1.823 & 16.371 & 1.609 & 0.623 & 3.07 & 3.98 & \textbf{2.87} \\ \cmidrule(l){2-16} 
    & ScopeFlow\cite{bar2020scopeflow} & 4.098 & 1.999 & 21.214 & 4.028 & 1.689 & 1.180 & 0.725 & 2.589 & 24.477 & 3.592 & 1.400 & 4.45 & 4.49 & 4.44 \\ \midrule
    
    \multirow{1}{*}{\begin{tabular}[c]{@{}l@{}} \emph{VM \vspace{5mm}} \end{tabular}} 
    & SfM-PM\cite{sfmPM2018structure} & 5.466 & 2.683 & \textbf{28.147} & 4.963 & 2.186 & 1.782 & 1.031 & 3.182 & 32.991 & \textbf{2.910} & 1.016 & \textbf{9.30} & 19.94 & \textbf{6.94} \\ \cmidrule(l){2-16} 
    & RicFlow \cite{hu2017robust}& 5.620 & 2.765 & 28.907 & 5.146 & 2.366 & 1.679 & 1.088 & 3.364 & 33.573 & 3.550 & 1.264 & 10.29 & 14.88 & 9.27 \\ \cmidrule(l){2-16} 
    & CPM \cite{hu2016efficient}& 5.960 & 2.990 & 30.177 & 5.038 & 2.419 & 2.143 & 1.155 & 3.755 & 35.136 & 3.557 & 1.189 & 13.85 & 18.71 & 12.77 \\ \cmidrule(l){2-16} 
    & CPM2 \cite{li2017coarse}& 6.180 & 3.012 & 32.008 & 5.059 & 2.399 & 2.126 & 1.212 & 3.625 & 37.014 & 3.253 & 0.980 & -  & -  & - \\ \cmidrule(l){2-16} 
    & EpicFlow\cite{revaud2015epicflow} & 6.285 & 3.060 & 32.564 & 5.205 & 2.611 & 2.216 & 1.135 & 3.727 & 38.021 & 4.115 & 1.360 & 16.69 & 24.34 & 15.00 \\ \cmidrule(l){2-16} 
    & \textbf{HybridFlow(SIFT)} & 8.082 & 4.966 & 33.445 & 7.513 & 4.907 & 3.983 & 2.635 & 5.401 & 41.585 & 7.018 & 4.086 & 23.57 & 19.19 & 24.32 \\ \cmidrule(l){2-16} 
    & \textbf{HybridFlow(DeepLab)} & 7.677 & 4.507 & 33.471 & 7.281 & 4.592 & 3.279 & 2.214 & 5.043 & 41.139 & 5.788 & 2.815 & 18.55 & 14.36 & 19.28 \\ \cmidrule(l){2-16} 
    & \textbf{HybridFlow} & \textbf{5.121} & \textbf{1.999} & 30.531 & \textbf{4.087} & \textbf{1.598} & \textbf{1.213} & \textbf{0.871} & \textbf{2.483} & \textbf{32.559} & 3.791 & \textbf{0.962} & 16.96 & \textbf{14.18} & 16.54 \\ \bottomrule
    \end{tabular}
}%
\end{table}

\begin{figure}[!ht]
\centering

    \begin{subfigure}[t]{0.2\textwidth}
		\includegraphics[width=\textwidth]{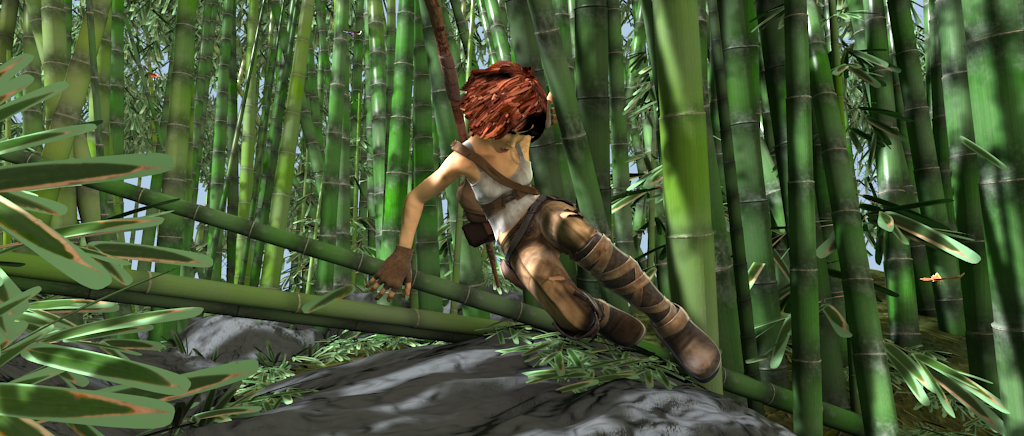}
		\label{fig:bamboo_first_frame}
	\end{subfigure}%
    \begin{subfigure}[t]{0.2\textwidth}
		\includegraphics[width=\textwidth]{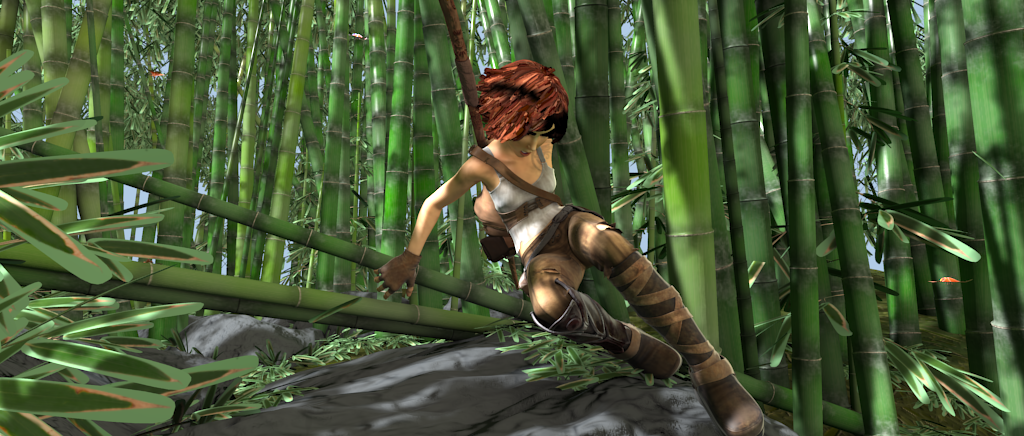}
		\label{fig:bamboo_second_frame}
	\end{subfigure}
	\begin{subfigure}[t]{0.2\textwidth}
		\includegraphics[width=\textwidth]{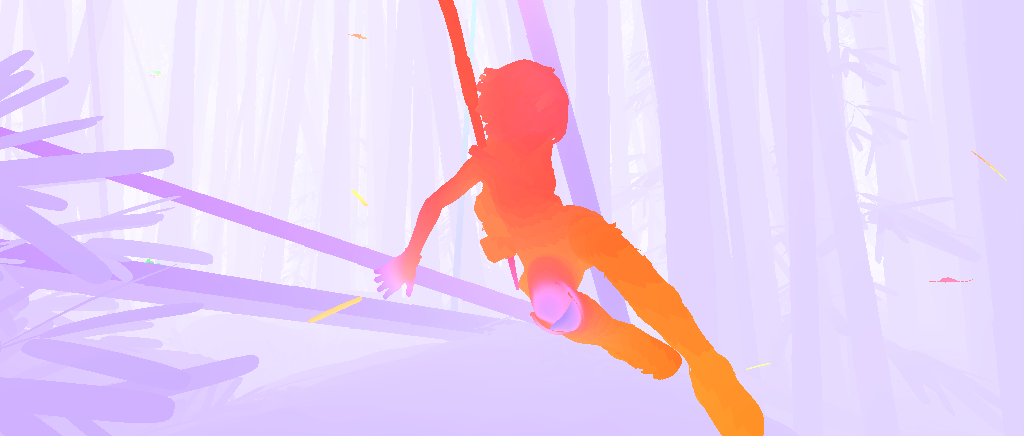}
		\label{fig:bamboo_True_flo}
	\end{subfigure}
    
    \begin{subfigure}[t]{0.4\textwidth}
		\includegraphics[width=\textwidth]{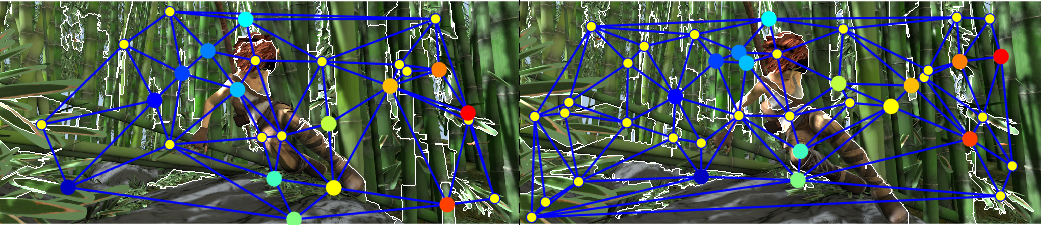}
		\label{fig:bamboo_FAZ_layer0}
	\end{subfigure}
	\begin{subfigure}[t]{0.2\textwidth}
		\includegraphics[width=\textwidth]{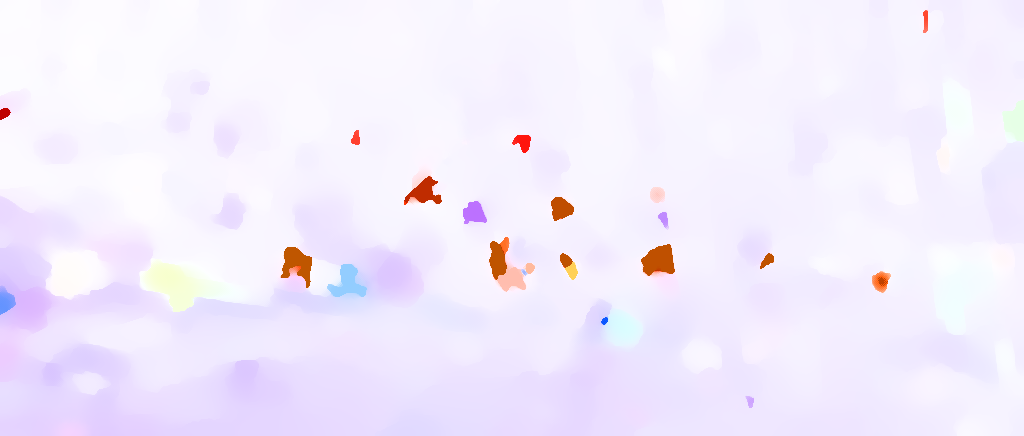}
		\label{fig:bamboo_FAZ_flo}
	\end{subfigure}
	
	\begin{subfigure}[t]{0.4\textwidth}
		\includegraphics[width=\textwidth]{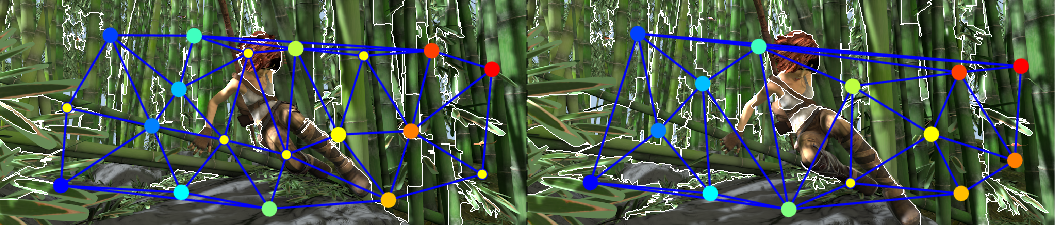}
		\label{fig:bamboo_slic_layer0}
	\end{subfigure}
	\begin{subfigure}[t]{0.2\textwidth}
		\includegraphics[width=\textwidth]{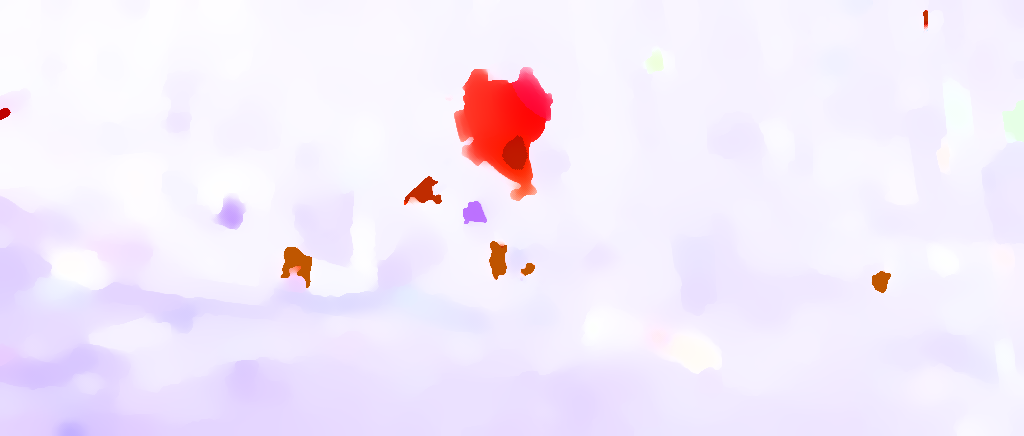}
		\label{fig:bamboo_slic_flo}
	\end{subfigure}
	
	\begin{subfigure}[t]{0.4\textwidth}
		\includegraphics[width=\textwidth]{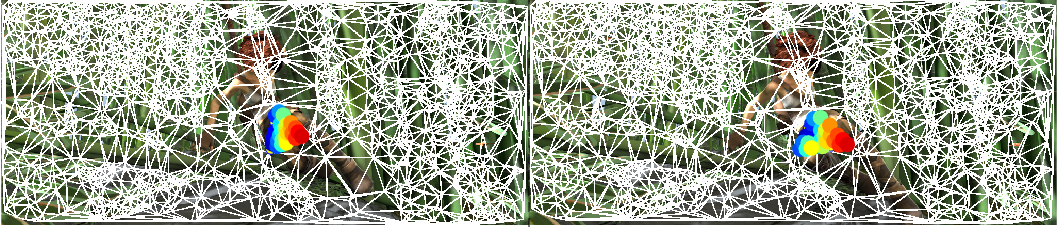}
		\label{fig:bamboo_DEL_layer0}
	\end{subfigure}
	\begin{subfigure}[t]{0.2\textwidth}
		\includegraphics[width=\textwidth]{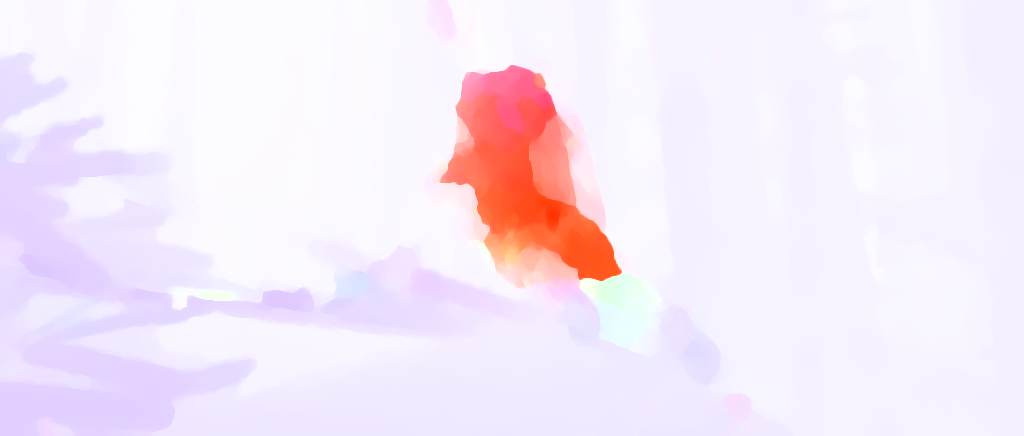}
		\label{fig:bamboo_DEL_flo}
	\end{subfigure}
	
	\begin{subfigure}[t]{0.4\textwidth}
		\includegraphics[width=\textwidth]{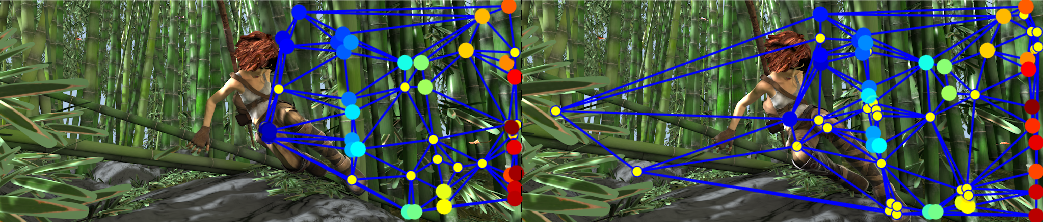}
		\label{fig:bamboo_cluster_layer0}
	\end{subfigure}
	\begin{subfigure}[t]{0.2\textwidth}
		\includegraphics[width=\textwidth]{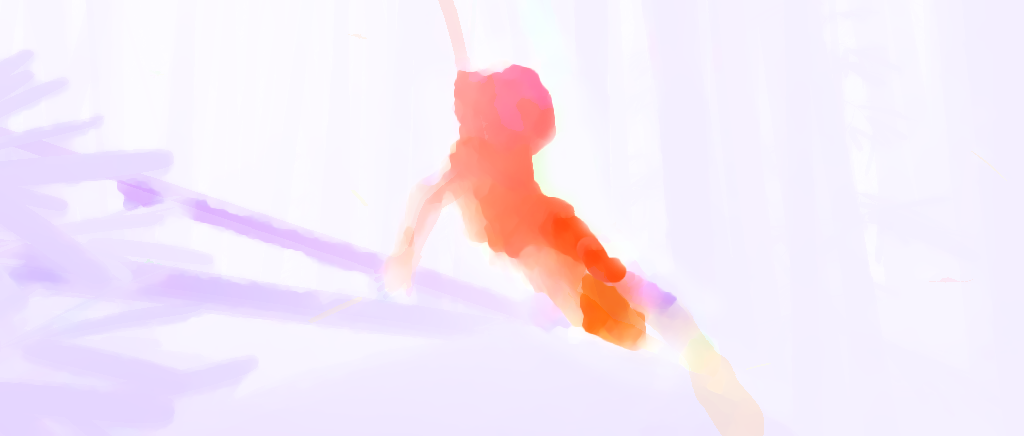}
		\label{fig:bamboo_cluster_flo}
	\end{subfigure}
	\caption{Graph Matching with different initial coarse-scale clustering methods on the pair of images shown in the 1st row. The initial coarse-scale clusters are resulting from Felsenszwalb's\cite{Felzenszwalb2004} graph-based segmentation (2nd row), SLIC superpixels \cite{achanta2012slic} (3rd row), Delaunay triangulation of rootSIFT features (4th row) and clustering of feature descriptors(5th row). The last column shows the optical flow results corresponding to each technique; ground truth shown in the 1st row.}
	\label{fig:initial_cluster_compare}
\end{figure}
\setlength{\textfloatsep}{10pt plus 1.0pt minus 2.0pt}

\subsection{Comparison of clustering techniques}
We compared initial coarse-scale clusters formed by (a) Delaunay triangulation of rootSIFT features, (b) SLIC superpixels, (c) Felsenszwalb's \cite{Felzenszwalb2004} graph-based image segmentation technique, and (d) our proposed clustering of feature descriptors. As shown in Figure \ref{fig:initial_cluster_compare}, initial coarse-scale clustering using SLIC, Felsenszwalb's graph-based technique and Delaunay triangulation of rootSIFT features cause erroneous results in graph matching, which accumulate in the finer-scales. However, coarse-scale clusters based on clustering feature descriptors provide consistent and robust performance. The average endpoint error (EPE) for the Sintel images in Figure \ref{fig:initial_cluster_compare} are 2.33, 2.12, 1.95 and 1.08 respectively. The last column shows the ground truth and below the resulting optical flow using each technique.
\subsection{Quantitative Evaluations}
\textbf{On Synthetic Data (MPI-Sintel)}
Table \ref{tab:table_results} shows the average endpoint error (EPE) on the MPI-Sintel ‘clean’ and  ‘final’ (realistic rendering effect) image dataset for HybridFlow and other state-of-the-art variational optical flow techniques. We present our results using three types of pixel-wise descriptors: (i) rootSIFT descriptors, named as HybridFlow(SIFT), (ii) features descriptors extracted from a pre-trained ResNet\cite{DBLP:ResNet101} trained on Sintel, named as HybridFlow(DeepLab), and (iii) descriptors learned by feature and context encoder as in RAFT\cite{raft2021-662}, name as HybridFlow. HybridFlow outperforms all other state-of-the-art variational techniques and gives comparable results to the deep-learning-based techniques with an average overall EPE of 5.121 in MPI-Sintel ‘final’ datasets.

\noindent
\textbf{On Real Data (KITTI-2015).} Table \ref{tab:table_results} shows the results for HybridFlow and other non-stereo-based optical flow methods on the 200 KITTI-2015 test images. Although HybridFlow does not have the best overall performance, it outperforms all variational techniques on the non-occluded test-set and has comparable performance for the other categories. Specifically, the percentage of background, foreground, and overall outliers are 31.06\%, 17.25\%, and 29.27\%, respectively. The percentages of outliers for non-occluded areas are 16.96\%, 14.18\%, and 16.54\%.

\subsubsection{Failure cases}

Graph matching is robust to texture variations, illumination variations, and deformations. However, erroneous matches can be introduced when large occluded areas fall inside the convex graph, as shown in the example in Figure \ref{fig:graph_matching_occlussion_example}. Mismatches in the graph matching can lead to the wrong matching of the finer-scale superpixels, and consequently, significant errors in the optical flow. This is clearly evident from the results in Table \ref{tab:table_results} for Sintel and KITTI-2015, where for the non-occluded test-sets, HybridFlow outperforms all state-of-the-art variational methods and matches the performance of deep-learning techniques such as ScopeFlow.

\section{Application: Large-scale 3D reconstruction}
The motivation of our work is large-scale 3D reconstructions from airborne images. In particular, we focus on full-motion video (FMV) and wide-area motion imagery, typically captured by a UAV/helicopter and an airplane, respectively. Deep learning techniques are not applicable since they have a fixed input size. Thus, a very high-resolution image must be scaled-down to typically less than $1K \times 1K$ to be used as input to the network. This significant reduction in resolution leads to low-resolution optical flow and significantly low fidelity 3D models. Most notably, there is no ground truth dataset for real scenarios to train the deep learning models.
On the other hand, the state-of-the-art variational methods considered in this work also impose restrictions on the input image size. For example, RicFlow and EpicFlow use a hierarchical structure employed by DeepMatching, which on an 8GB GPU can only handle $1K \times 1K$ resolutions. HybridFlow can handle arbitrary-sized resolutions with a low memory footprint. In this section, we present the results of the application of HybridFlow on the use case of large-scale 3D reconstruction from airborne images. We reiterate that there is no ground truth data for training models in such scenarios, and the resolutions can be significantly higher than $1K \times 1K$. 

\subsection{Image-based Large-scale Reconstruction}
Image-based reconstruction involves three main components: (1) Structure-from-Motion (SfM) for camera pose estimation, (2) Bundle Adjustment optimization, and (3) Multi-View Stereo (MVS). In contrast, we reformulate the reconstruction as a single-step process. Using HybridFlow allows us to triangulate directly the dense matches without MVS as a post-processing step, therefore achieving faster reconstructions.

\begin{figure}[!ht]
\centering
\includegraphics[width=0.49\textwidth]{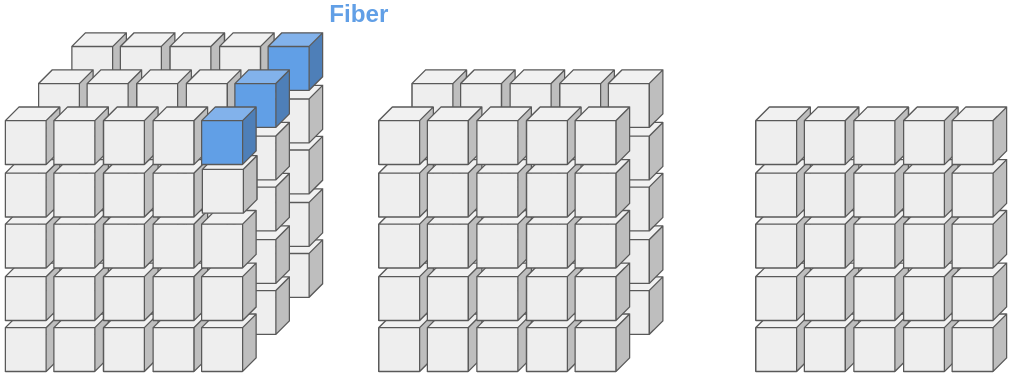}
\caption{On-disk dynamic tensor-shaped data structure. For each image, we store a tensor with layers containing pixel-level matches to subsequent images based on the HybridFlow. Unmatched pixels in the second image are stored in the tensor data structure for the second image, which contains layers with pixel-level matches to the third image and onward. A fiber is shown in blue. Each cell contains the match of that pixel, i.e. the top right corner in all subsequent images. Reconstruction is reduced to triangulating the matches contained within each fiber.}
\label{fig:tensor}
\end{figure}


We design a specialized off-memory, on-disk data structure for storing the matches. As shown in Figure \ref{fig:tensor}, at every image, we keep a tensor with layers containing pixel-level matches to subsequent images based on the HybridFlow. Unmatched pixels in the second image are stored in the tensor data structure for the second image, which contains layers with pixel-level matches to the third image and onwards. The data structure can scale up dynamically to arbitrary-sized datasets (subject to the disk limits) and allows for efficient outlier removal and validation, i.e. multiple pixels in the same image cannot be matched to the same pixel in the following image. A simple look-up at a fiber of the tensor gives the matches for that pixel in all subsequent images. Hence, reconstruction is reduced to traversing all fibers in each tensor and triangulating to get a 3D position.

We demonstrate the effectiveness of HybridFlow on large-scale reconstruction from images and present result on two different types of datasets: full-motion video, and wide-area motion imagery. We followed the single step process described above employing the dynamic tensor-shaped data structure for the efficient processing of the matches calculated by HybridFlow.

\begin{table*}[!t]
\resizebox{0.969\textwidth}{!}{
\begin{tabular}{lllllllllll}
\hline
\textbf{Method} & \textbf{Features} & \textbf{\# matches} & \textbf{SfM Variant} & \textbf{\begin{tabular}[c]{@{}l@{}}time-SfM\\ (min)\end{tabular}} & \textbf{\# points} & \textbf{MVS Variant} & \textbf{\# points} & \textbf{\begin{tabular}[c]{@{}l@{}}time-MVS\\ (min)\end{tabular}} & \textbf{Run-time} & \textbf{\begin{tabular}[c]{@{}l@{}}Reprojection\\ Error\end{tabular}} \\ \hline
VisualSFM\cite{wu2011multicore} & SIFT\cite{lowe2004sift} & 35,0161 & Bundler\cite{snavely2006photo} & 1.10 & 4,863 & PMVS\cite{furukawa2007accurate} & 111,189 & 1.289 & 2.389 & 0.880 \\ \hline
COLMAP\cite{schonberger2016structure} & RootSIFT\cite{ArandjelovicZisserman:2012} & 704,127 & iterative BA & 11.051 & 11,274 & Kazhdan\cite{schonberger2016pixelwise} & 287,205 & 23.965 & 35.016 & 0.810 \\ \hline
Our method & Epic-Flow\cite{revaud2015epicflow} & 1,576,705 & iterative BA & 12.533 & 139,606 & - & 139,606 & - & 12.533 & 1.004 \\ \hline
Our method & HybridFlow & 8,144,093 & iterative BA & 16.052 & 6,512,324 & - & 6,512,324 & - & 16.052  & 0.820 \\ \hline
\end{tabular}%
}
\caption{The comparison of number of points reconstructed and reprojection error.}
\label{tabel:3d_pointcloud_numbers}
\end{table*}

\vspace{-10pt}
\subsection{Full-motion Video}
Full-motion video(FMV) is typically captured by a helicopter at an oblique aerial angle so that the rooftops and the facades of the buildings are visible in the images. The ground sampling density is significantly higher than that of a satellite image, i.e. in the order of a few cms, and can vary according to the aircraft’s flight height, depending on the area it is flying over.

We ran experiments on a full-motion video dataset containing images taken from a helicopter circling an area containing a few mockup buildings. Our test dataset contains 71 images with resolution $1280  \times 720$ with unknown camera calibrations or EXIF information. We report results using the (i) single-step reconstruction using HybridFlow matches, the (ii) same single-step reconstruction using EpicFlow matches, (iii) and the state-of-the-art incremental SfM techniques Bundler\cite{snavely2006photo}, VisualSFM\cite{wu2011multicore}, COLMAP\cite{schonberger2016structure}. 

\begin{figure}[!ht]
\centering
    \begin{subfigure}[t]{0.22\textwidth}
		\centering	
		\includegraphics[width=\textwidth]{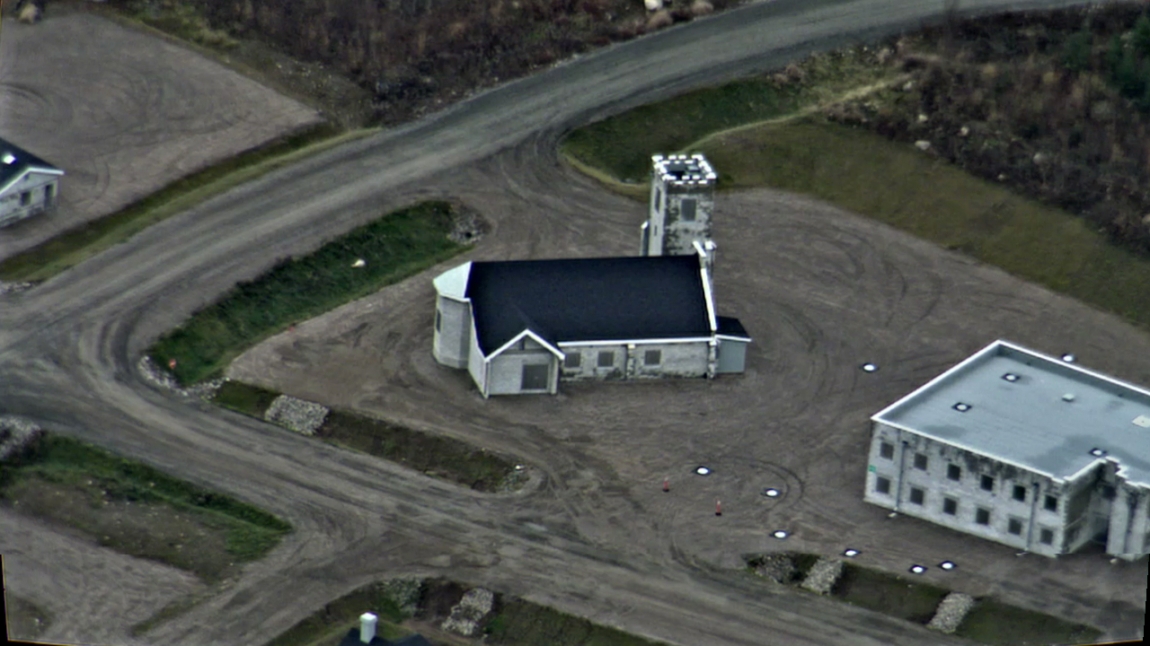}
		\caption{}
		\label{fig:church001}
	\end{subfigure}
	\begin{subfigure}[t]{0.22\textwidth}
		\centering	
		\includegraphics[width=\textwidth]{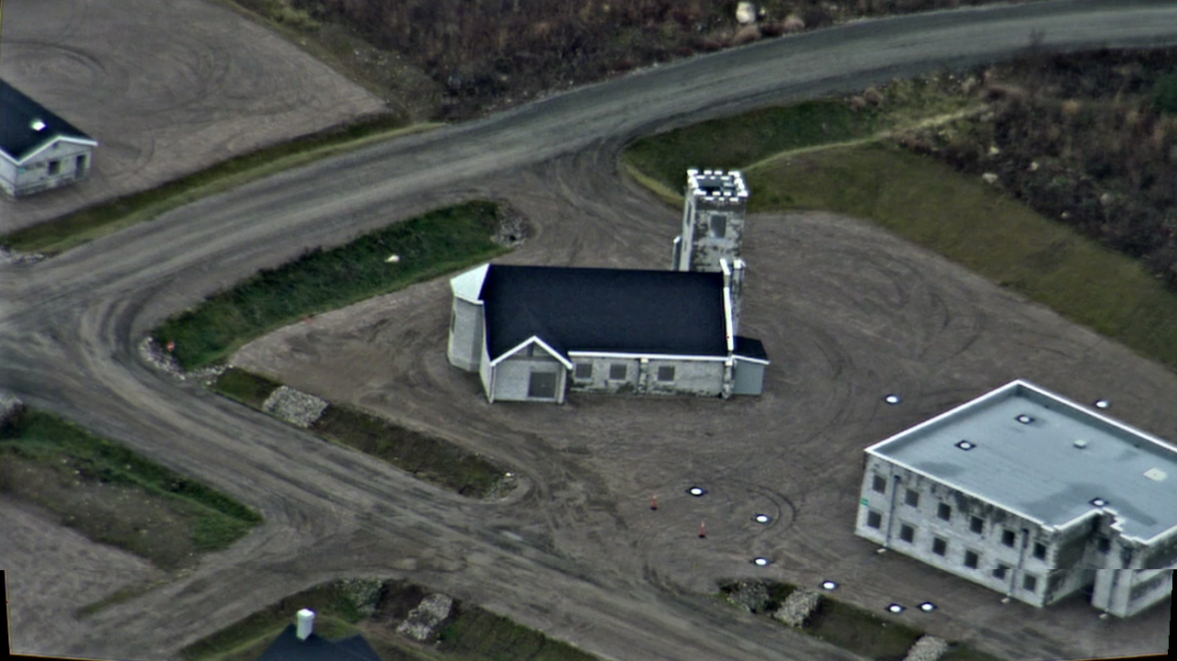}
		\caption{}
		\label{fig:church002}
	\end{subfigure}
	\begin{subfigure}[t]{0.44\textwidth}
		\centering	
		\includegraphics[width=\textwidth]{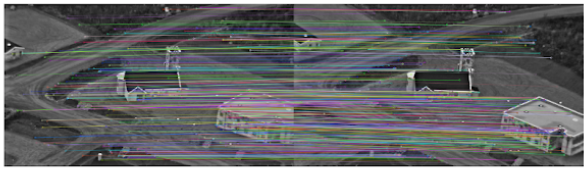}
		\caption{}
		\label{fig:sift_matches_church}
	\end{subfigure}
	
	\begin{subfigure}[t]{0.44\textwidth}
		\centering	
		\includegraphics[width=\textwidth]{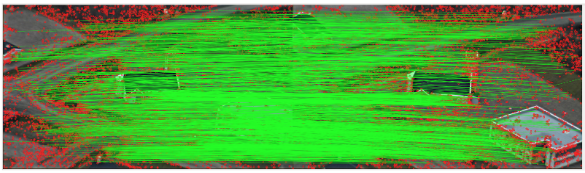}
		\caption{}
		\label{fig:rootSIFT_matches_church}
	\end{subfigure}
	\begin{subfigure}[t]{0.22\textwidth}
		\centering	
		\includegraphics[width=\textwidth]{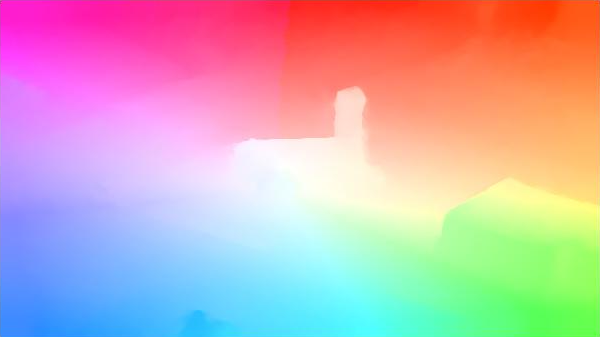}
		\caption{}
		\label{fig:Epicflow}
	\end{subfigure}
	\begin{subfigure}[t]{0.22\textwidth}
		\centering	
		\includegraphics[width=\textwidth]{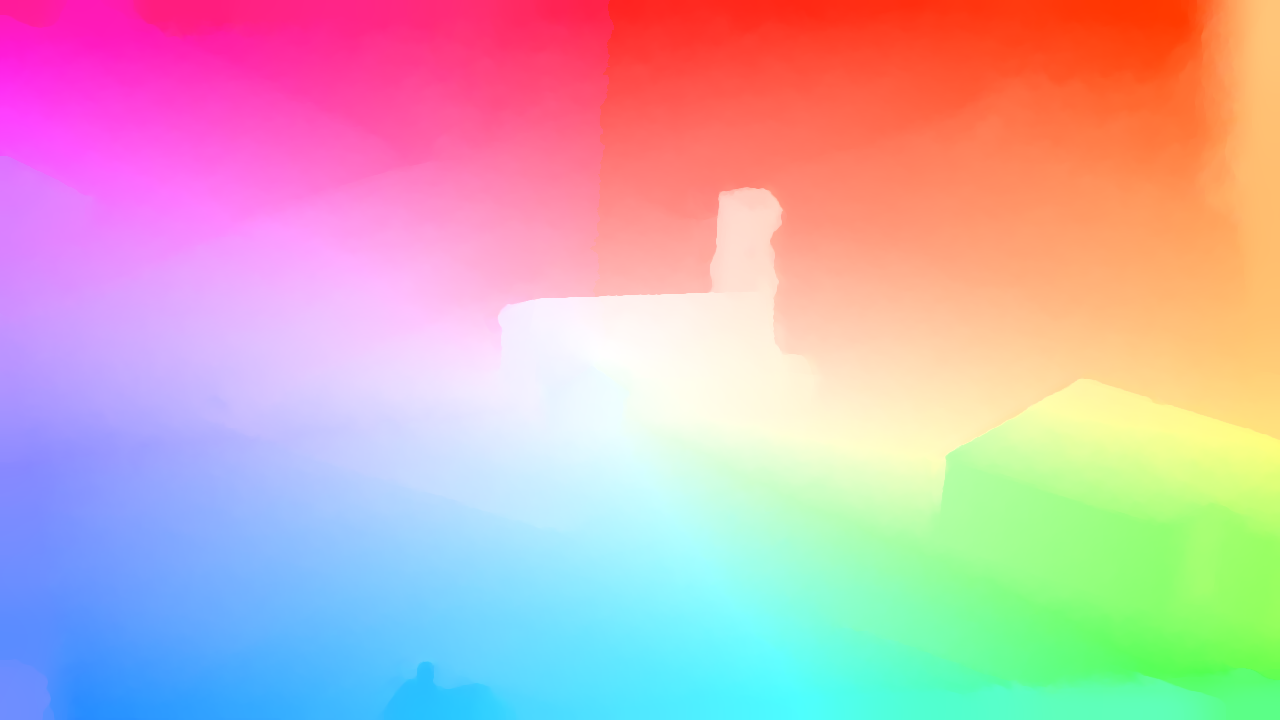}
		\caption{}
		\label{fig:Hybridflow}
	\end{subfigure}
    \caption{Density of matches. The first row (a) and (b) shows an example of the input image frames, (c) shows SIFT\cite{lowe2004sift} matches, (d) shows RootSIFT\cite{ArandjelovicZisserman:2012} matches, (e) and (f) shows EpicFlow\cite{revaud2015epicflow} and HybridFlow results.}	
    \label{fig:matches_compare}
\end{figure}

Perhaps the most popular feature extraction methods used in SfM is SIFT\cite{lowe2004sift}. In COLMAP\cite{schonberger2016structure}, they use a modified version called RootSIFT\cite{ArandjelovicZisserman:2012} for extracting and matching each image. The first comparison focuses on the density of the matches. Figure \ref{fig:sift_matches_church} shows the SIFT matches, Figure \ref{fig:rootSIFT_matches_church} the RootSIFT matches, Figure \ref{fig:Epicflow} the EpicFlow matches, and Figure \ref{fig:Hybridflow} the HybridFlow matches for the input images shown in Figures \ref{fig:church001} and \ref{fig:church002}. The latter two show the matches as colour-coded optical flows for visualization clarity, otherwise drawing the matches will cover the entire image. Table \ref{tabel:3d_pointcloud_numbers} presents the total number of matches per technique. As expected, SIFT and RootSIFT have the lowest number of matches since they only extract scale-space extrema. On the other hand, the dense optical flow technique EpicFlow results in eight times lower number of matches than HybridFlow. 

\begin{figure*}[!ht]
\centering
    \begin{subfigure}[t]{0.24\textwidth}
		\centering	
		\includegraphics[width=\textwidth]{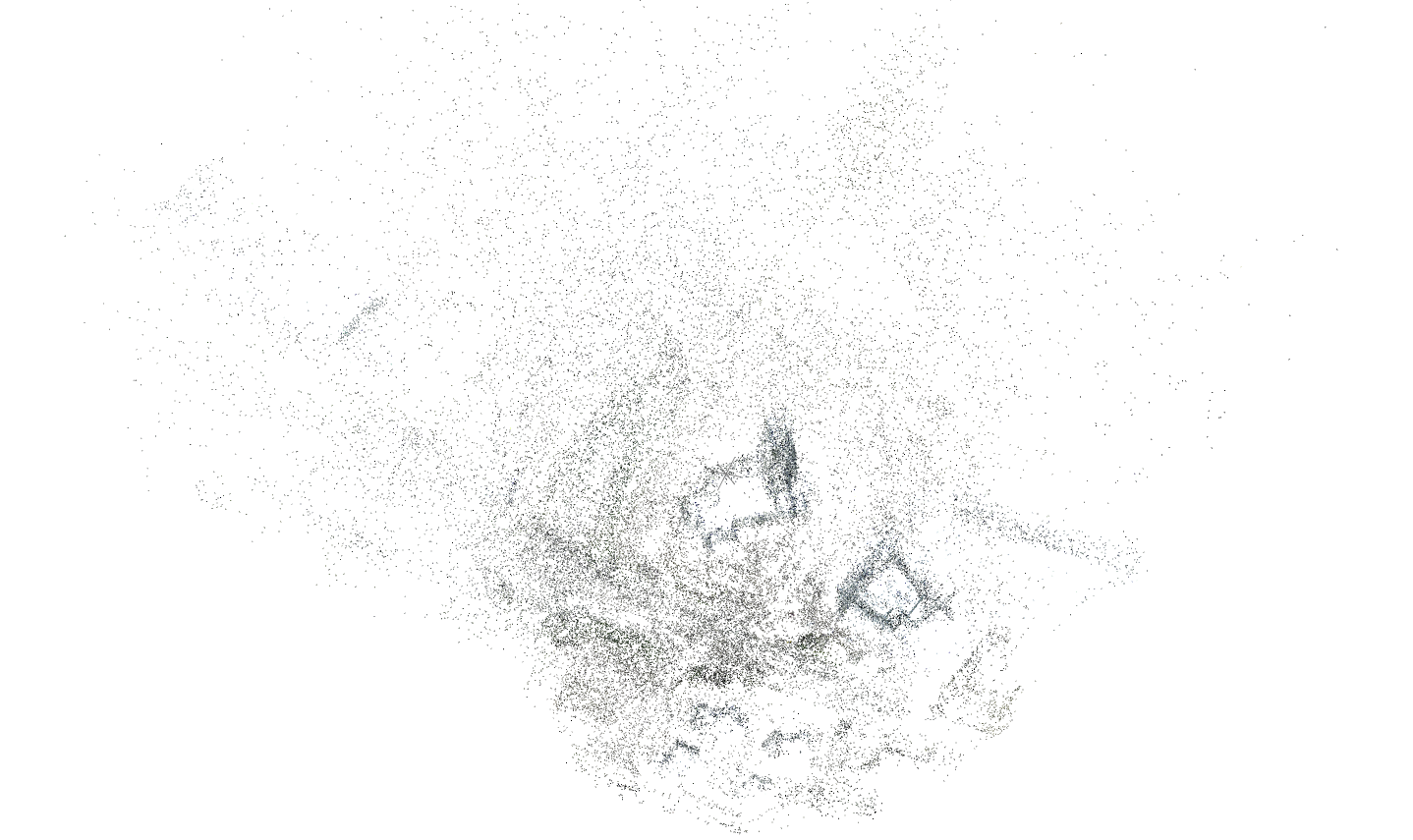}
		\caption{}
		\label{fig:colmap_sparse}
	\end{subfigure}
	\begin{subfigure}[t]{0.24\textwidth}
		\centering	
		\includegraphics[width=\textwidth]{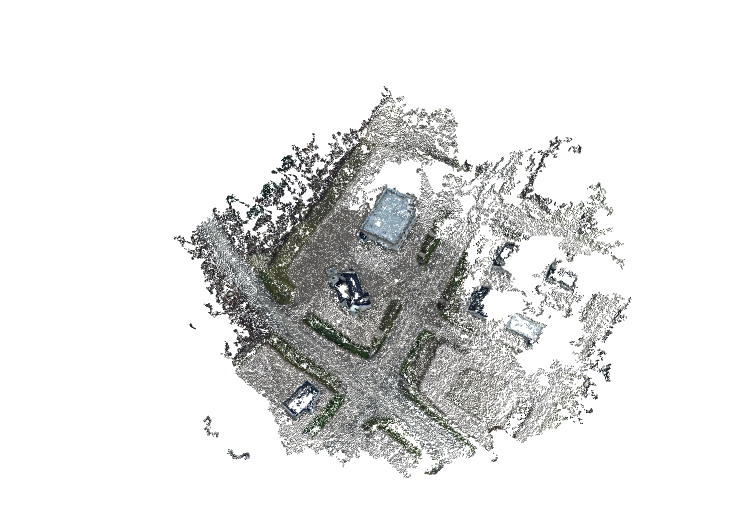}
		\caption{}
		\label{fig:colmap_dense}
	\end{subfigure}
	\begin{subfigure}[t]{0.24\textwidth}
		\centering	
		\includegraphics[width=\textwidth]{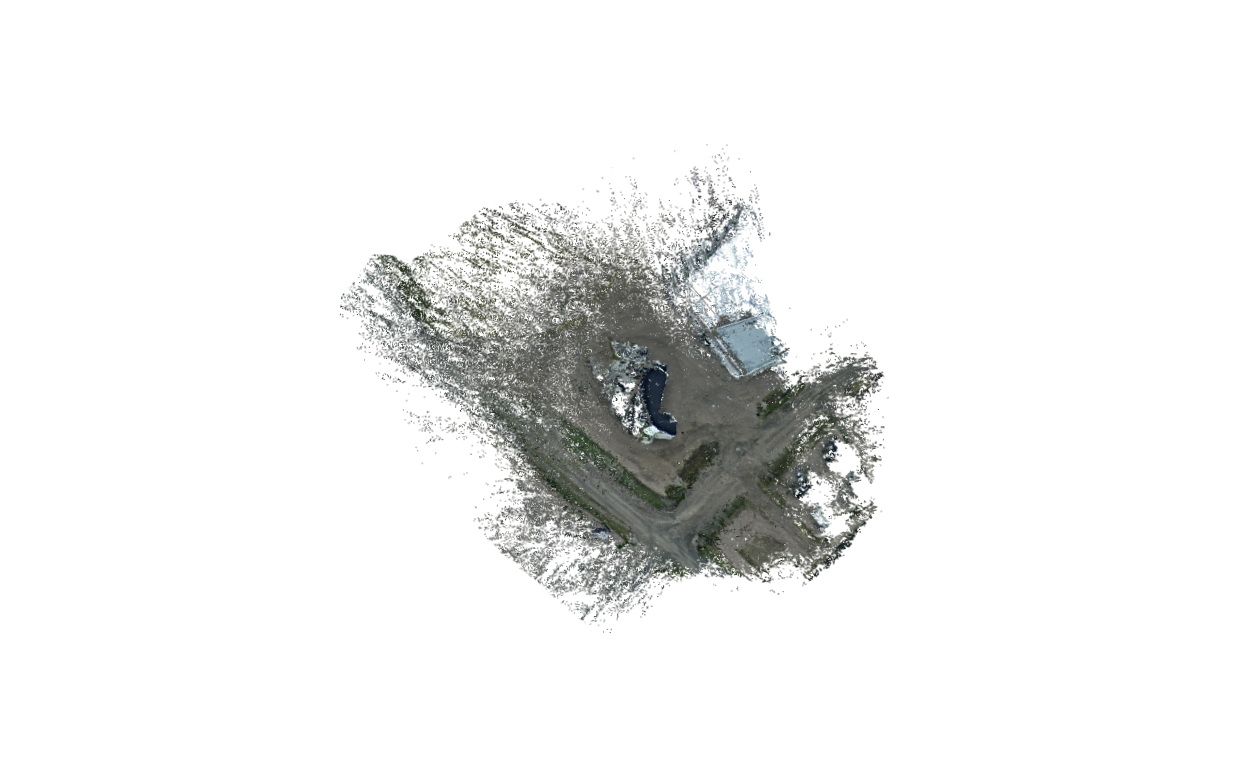}
		\caption{}
		\label{fig:epicflow_3d}
	\end{subfigure}
	\begin{subfigure}[t]{0.24\textwidth}
		\centering	
		\includegraphics[width=\textwidth]{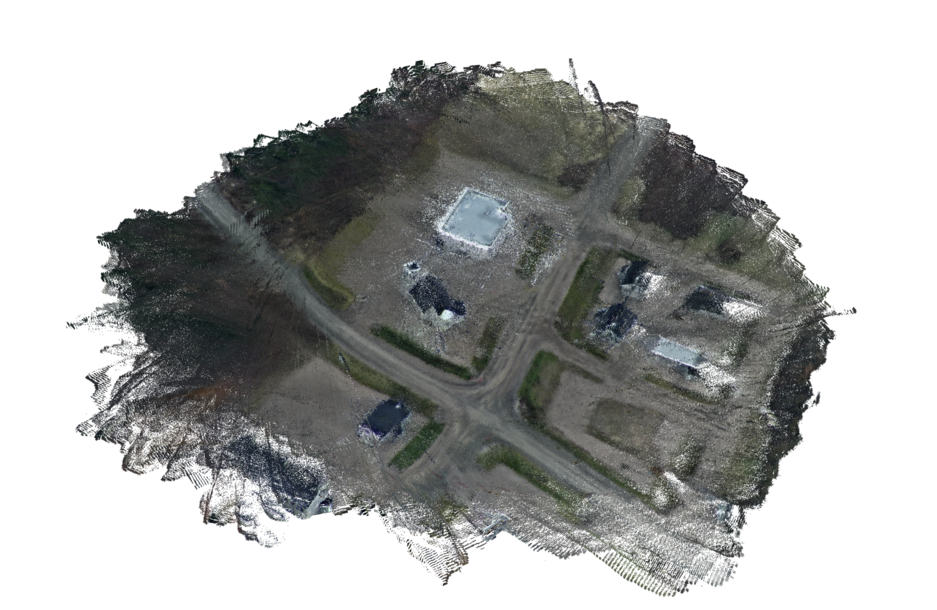}
		\caption{}
		\label{fig:hybridflow_3d}
	\end{subfigure}
    \caption{The reconstruction serves as a proxy to the accuracy of the matches. We calculate and compare reprojection errors for the techniques shown in Table \ref{tabel:3d_pointcloud_numbers}. (a) shows COLMAP's sparse (SfM) reconstruction, (b) COLMAP's dense (MVS) reconstruction \cite{schonberger2016structure}, (c) shows our single step reconstruction using dense matches from Epicflow \cite{revaud2015epicflow}, and (d) our single step reconstruction with Hybridflow. HybridFlow produces 60x more matches than COLMAP and 47x more matches than EpicFlow. The reprojection error is comparable with COLMAP (for 60x more points) while the runtime is less than half.}	
    \label{fig:pointcloud_compare}
\end{figure*}

The reconstruction can serve as a proxy for the accuracy of the matches in cases where ground truth is not available. We proceed with the evaluation of the reconstruction in terms of the reprojection error. Figure \ref{fig:pointcloud_compare} shows the reconstructed pointcloud of (a) COLMAP's sparse (SfM) reconstruction, (b) COLMAP's dense (MVS) reconstruction, (c) our single-step reconstruction using HybridFlow matches, and (d) our single-step reconstruction using EpicFlow matches. The reconstructed point clouds are rendered from the same viewpoint and camera intrinsics. The reprojection error using our single-step method with HybridFlow achieves the highest number of reconstructed points in the lowest time per point, while the reprojection error is comparable with COLMAP for almost 60x more points.


\vspace{-20pt}
\subsection{Wide-area Motion Imagery}
Wide-area motion imagery (WAMI) is captured by an aircraft flying at over 10,000ft and can cover areas of $10-20km^2$. The aircraft orbits around the area of interest during the flight, and an array of cameras captures and streams image data at about two frames per second.

\begin{figure*}[!ht]
\centering
\begin{subfigure}[t]{0.45\textwidth}
\centering
    	\centering	
		\includegraphics[width=\textwidth]{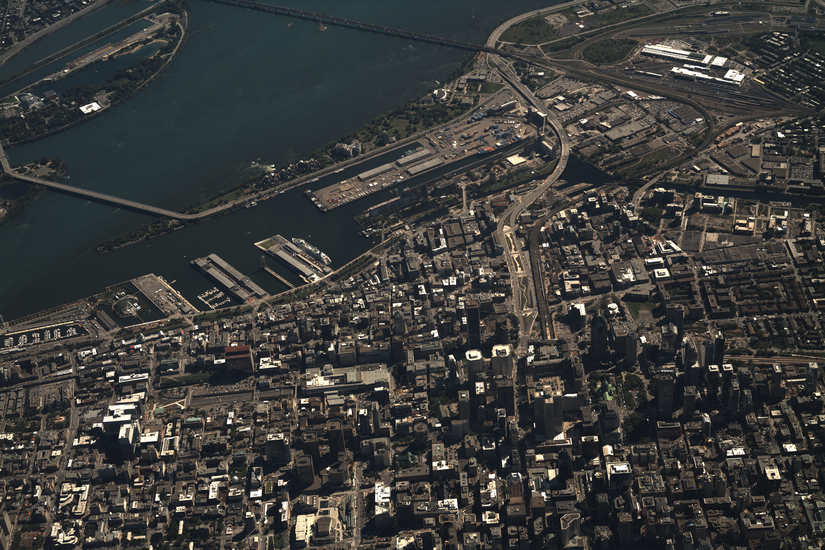}
		\caption{}
		\label{fig:Montreal_frame}
	\label{fig:Montreal_frame1}
\end{subfigure}
\begin{subfigure}[t]{0.45\textwidth}
\centering
		\includegraphics[width=\textwidth]{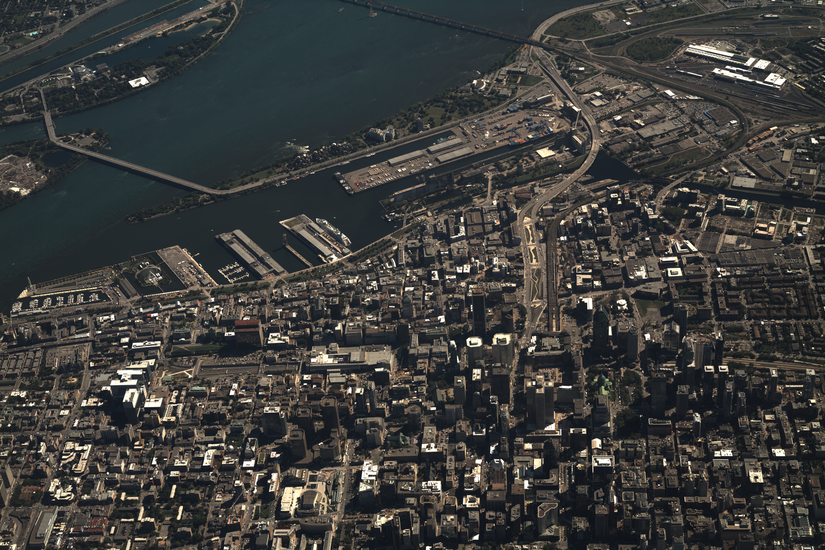}
		\caption{}
		\label{fig:Montreal_frame2}
\end{subfigure}

\begin{subfigure}[t]{0.45\textwidth}
\centering
		\includegraphics[width=\textwidth]{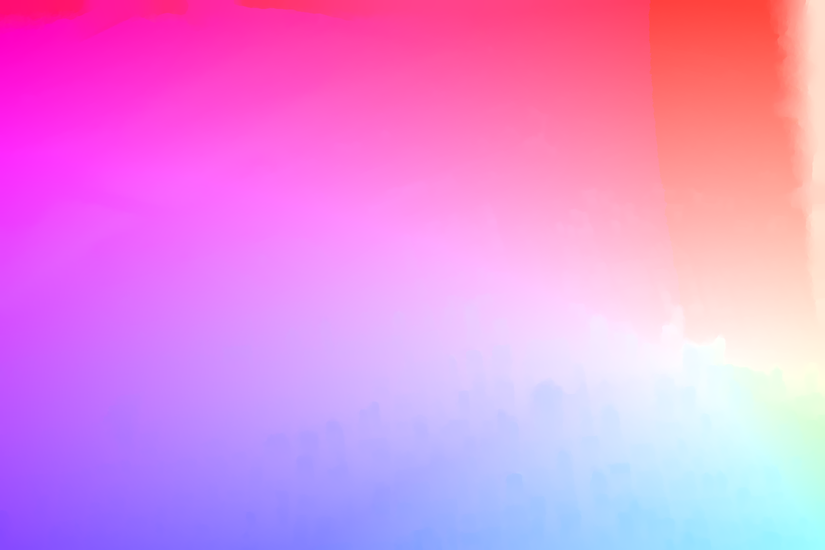}
		\caption{}
		\label{fig:montreal_flo}
\end{subfigure}
\begin{subfigure}[t]{0.45\textwidth}
\centering
		\includegraphics[width=\textwidth]{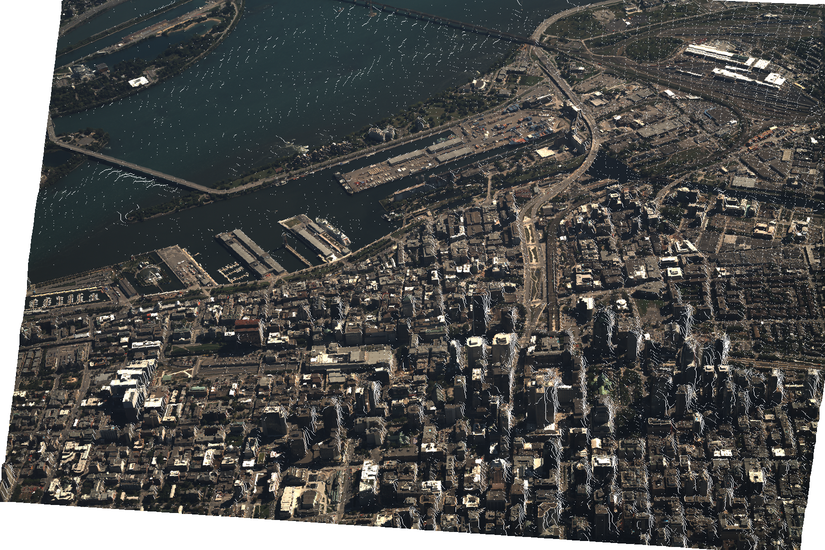}
		\caption{}
		\label{fig:montreal_recreated}
\end{subfigure}

\begin{subfigure}[t]{\textwidth}
\centering
		\includegraphics[width=0.9\textwidth]{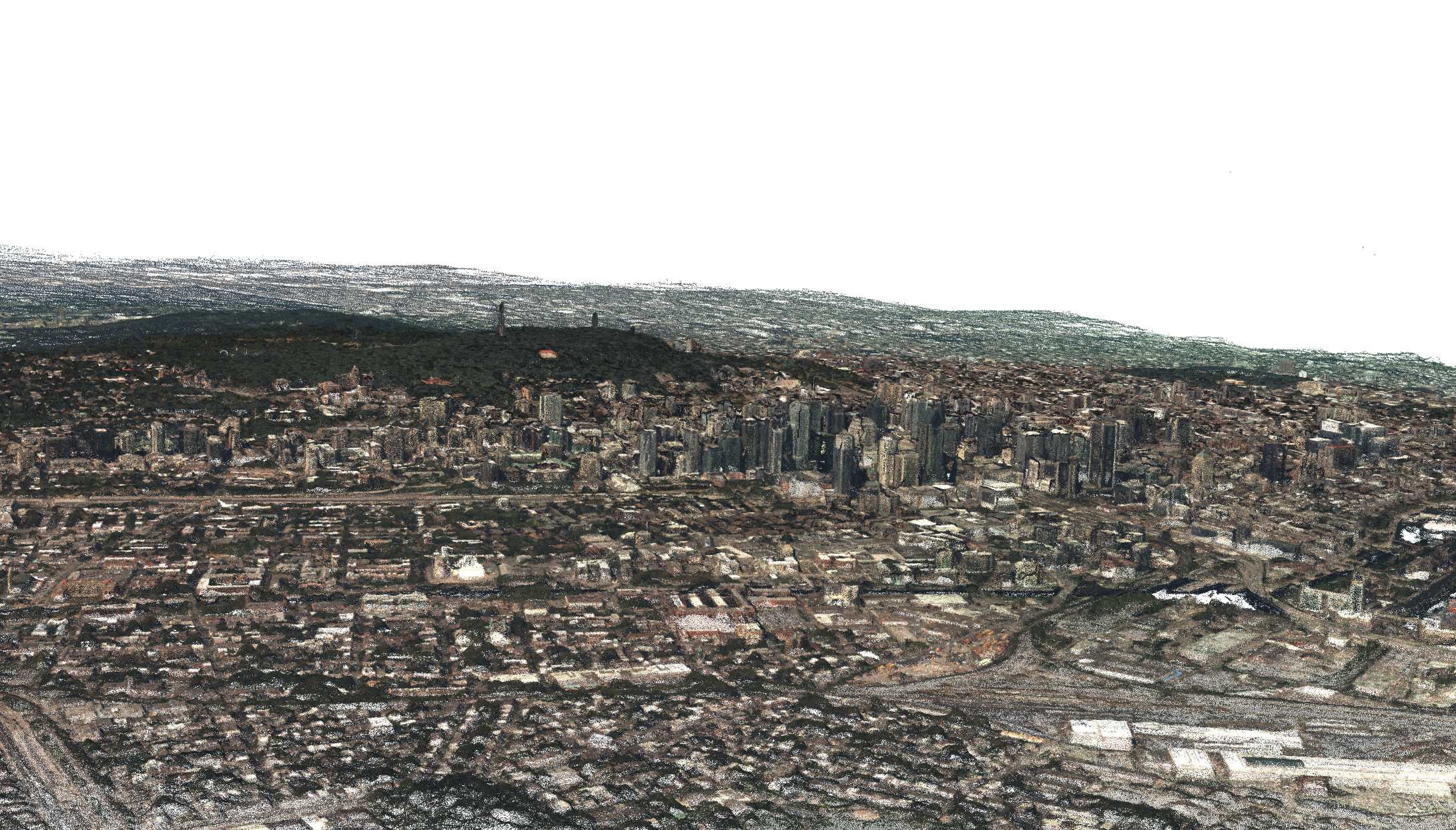}
		\caption{}
		\label{fig:montreal_pts}
\end{subfigure}
\caption{(a) and (b) are two consecutive WAMI images of a downtown urban area with resolution $6600\times 4400$. (c) HybridFlow is the only top-performing variational method that can handle high-resolution images. Deep learning techniques cannot be applied due to the fixed input size of the networks as explained in the text. (d) Image resampled from (a) using HybridFlow flows in (c) to form (b). (e) Reconstructed pointcloud using 320 images.}
\end{figure*}

Figure \ref{fig:Montreal_frame} shows an example of a WAMI image capturing a downtown urban area. The resolution is $6600\times 4400$ is considered average amongst WAMI, since some of the larger resolutions can reach sizes of up to $14000\times 12000$. Deep learning techniques can be applied only (i) by rescaling the image to the fixed input size expected by the neural network, or (ii) tiling the image, calculating flows per tile, and then merging the results. In the first case, rescaling reduces the resolution and subsequently the final number of reconstructed points. Furthermore, essential details such as cars and trees are completely removed. In the latter case, there is no one-to-one mapping between tiles. For example, a tile may contain areas appearing in two or more different tiles in the second image. Furthermore, the deep optical flow techniques always return a match for every pixel. That means that even if an area is not present in a tile, this will nevertheless be matched to another area in the second image. For these reasons, deep learning techniques cannot be applied in these use cases. 

Competing variational methods such as RicFlow \cite{hu2017robust}, EpicFlow \cite{revaud2015epicflow} cannot be applied either since hierarchical structure employed by DeepMatching \cite{revaud2016deepmatching}, which on an 8GB GPU can only handle $1K \times 1K$ resolutions. In contrast, HybridFlow is the only top-performing variational method that can handle arbitrary-sized images such as large WAMI. Figure \ref{fig:Montreal_frame} and \ref{fig:Montreal_frame2} shows two consecutive images capturing a downtown urban area having a resolution of $6600\times 4400$. HybridFlow is the only top-performing variational method that can handle high-resolution images as shown in Figure \ref{fig:montreal_flo}. Deep learning techniques cannot be applied due to the fixed input size of the networks. Similarly, competing state-of-the-art variational methods cannot be applied for this size of images as explained above. Figure \ref{fig:montreal_recreated} shows the resampled image from Figure \ref{fig:Montreal_frame2} using the HybridFlow matches in Figure \ref{fig:montreal_flo} and the matched pixels in Figure \ref{fig:Montreal_frame}. Figure \ref{fig:montreal_pts} shows a render of the reconstructed pointcloud for the downtown urban area generated using 320 images of the same size.

\clearpage

\section{Conclusion}
\label{sec:conclusion}
We addressed the problem of large displacement optical flow and presented a hybrid approach based on sparse feature matching using feature descriptors and graph matching, named HybridFlow. In contrast to state-of-the-art, it does not require training, and the use of sparse feature matching is robust and can scale up to arbitrary image sizes. This makes our technique applicable in use-cases such as reconstruction or object tracking where ground-truth is unavailable, and processing must be performed in interactive time. We match initial coarse-scale clusters based on a clustering of context features. We employ graph matching to match perceptual groups clustered using SLIC superpixels within each initial coarse-scale cluster, and perform pixel matching on smaller clusters. Based on the combined feature matches and the graph-node matches, we calculate the initial flow which is interpolated using an edge-preserving interpolation and refined using variational refinement. The proposed technique has been evaluated on two benchmark datasets (Sintel, KITTI), and we compared it with the current state-of-the-art variational optical flow techniques. We show that HybridFlow surpasses all other state-of-the-art variational methods in non-occluded test sets. Specifically, for Sintel, HybridFlow has the lowest overall EPE, while for KITTI, it gives comparable results.


\section*{Acknowledgment}
This research is based upon work supported by the Natural Sciences and Engineering Research Council of Canada Grants No. N01670 (Discovery Grant) and DNDPJ515556-17 (Collaborative Research and Development with the Department of National Defence Grant). 

\section*{Data availability statement}
The datasets generated and analysed during the current study are available online: Sintel \cite{Butler:ECCV:2012}, and the KITTI \cite{Menze2015CVPR} benchmark datasets.

\bibliography{main}

\end{document}